\documentclass[lettersize,journal]{IEEEtran}
\usepackage{amsmath,amsfonts}
\usepackage{algorithmic}
\usepackage{algorithm}
\usepackage{array}
\usepackage{textcomp}
\usepackage{stfloats}
\usepackage{url}
\usepackage{verbatim}
\usepackage{graphicx}

\usepackage{booktabs}
\usepackage{multirow}

\usepackage[margin=1in]{geometry}
\usepackage{amsmath}
\usepackage{color,soul}
\usepackage[dvipsnames]{xcolor}
\usepackage[numbers, sort&compress]{natbib}
\usepackage{caption}
\usepackage{subcaption}
\usepackage{makecell}

\usepackage{enumitem}
\setlist{leftmargin=8pt}
\setlist[itemize]{topsep=0pt, itemsep=0pt, parsep=0pt, labelsep=2pt}

\usepackage[hidelinks]{hyperref}
\usepackage{tikz}

\usepackage[framemethod=TikZ]{mdframed}
\usepackage{lipsum}
\definecolor{hlcolor}{RGB}{255,231,173}
\sethlcolor{hlcolor}

\mdfdefinestyle{LLMFrame}{%
    linecolor=blue,
    outerlinewidth=1pt,
    roundcorner=10pt,
    innertopmargin=4pt,
    innerbottommargin=4pt,
    innerrightmargin=5pt,
    innerleftmargin=5pt,
    backgroundcolor=white}
    
\usepackage{balance}
\newcommand{\RN}[1]{%
  \textup{\uppercase\expandafter{\romannumeral#1}}%
}

\definecolor{color1}{RGB}{97,127,178}

\begin{document}
%
\title{Virtual Roads, Smarter Safety: A Digital Twin Framework for Mixed Autonomous Traffic Safety Analysis}


\author{Hao Zhang, Ximin Yue, Kexin Tian, Sixu Li, Keshu Wu, Zihao Li, Dominique Lord, Yang Zhou* 
\thanks{*Corresponding author. (Email: yangzhou295@tamu.edu).}
\thanks{H. Zhang, X. Yue, K. Tian, S. Li, Z. Li, D. Lord, Y. Zhou are associated with the Zachry Department of Civil and Environmental Engineering, Texas A\&M University, College Station, TX, 77840, USA.}
\thanks{K. Wu is jointly affiliated with the Department of Landscape Architecture and Urban Planning, and the Zachry Department of Civil and Environmental Engineering, Texas A\&M University, College Station, TX, 77840, USA.} 
}

%



\IEEEtitleabstractindextext{%
\begin{abstract}
This paper presents a digital-twin platform for active safety analysis in mixed traffic environments. The platform is built using a multi-modal data-enabled traffic environment constructed from drone-based aerial LiDAR, OpenStreetMap, and vehicle sensor data (e.g., GPS and inclinometer readings). High-resolution 3D road geometries are generated through AI-powered semantic segmentation and georeferencing of aerial LiDAR data. To simulate real-world driving scenarios, the platform integrates the CAR Learning to Act (CARLA) simulator, Simulation of Urban MObility (SUMO) traffic model, and NVIDIA PhysX vehicle dynamics engine. CARLA provides detailed micro-level sensor and perception data, while SUMO manages macro-level traffic flow. NVIDIA PhysX enables accurate modeling of vehicle behaviors under diverse conditions, accounting for mass distribution, tire friction, and center of mass. This integrated system supports high-fidelity simulations that capture the complex interactions between autonomous and conventional vehicles. Experimental results demonstrate the platform’s ability to reproduce realistic vehicle dynamics and traffic scenarios, enhancing the analysis of active safety measures. Overall, the proposed framework advances traffic safety research by enabling in-depth, physics-informed evaluation of vehicle behavior in dynamic and heterogeneous traffic environments.
\end{abstract}

}

\maketitle

\IEEEdisplaynontitleabstractindextext

%
\IEEEpeerreviewmaketitle

\section{Introduction}
%
%
%
%
\IEEEPARstart{W}{ith} rising complexity in urban mobility and the increasing prevalence of mixed traffic environments, accurately analyzing active safety scenarios becomes imperative \cite{li2024beyond,li2024disturbances,zhang2024anticipatory}. As the transportation ecosystem evolves with the integration of automated vehicles (AVs), vulnerable road users (VRUs), and increasingly complex roadway geometries, the need for high-fidelity, dynamic simulation frameworks have never been more critical \cite{pauwels2022safety}. The emergence of digital-twin technology offers revolutionary potential in this area, providing virtual replicas of physical environments that facilitate comprehensive analysis and proactive interventions \cite{almeaibed2021digital,kuvsic2023digital, wang2021digital}.

Traditional passive safety analysis methods, which rely on historical crash data and retrospective accident studies, face serious limitations in today's data-scarce and technologically evolving environments \cite{mahmud2016traditional}. These methods are inherently reactive, capturing the consequences of crashes rather than the precursors, and thus fall short in addressing the proactive safety demands of emerging mobility systems \cite{lord2010statistical}. The rarity of severe crashes further exacerbates this issue, particularly in underserved communities or in the context of novel technologies such as AVs and connected automated vehicles (CAVs), where empirical data remains limited and inconsistent \cite{khan2024advancing}.Even when crash data is available, it often suffers from reliability issues—such as underreporting, missing behavioral context, and lack of ground truth—that limit its usefulness for future safety assessment \cite{useche2019more}. For instance, crash databases typically lack precise contextual information such as vehicle trajectories, sensor failures, or interaction dynamics, which are crucial for understanding causality in AV-related incidents \cite{favaro2017examining}. The black-box nature of crash causality and the lack of interpretability further constrain our ability to extract meaningful insights from passive safety data \cite{goudarzi2024collision}.
 
To address these challenges, active safety analysis has emerged as a proactive approach that continuously monitors vehicle states, environmental conditions, and traffic interactions to detect and mitigate potential hazards before they escalate into collisions \cite{astarita2019surrogate, li2025adaptive}. However, existing methods often fall short due to oversimplified assumptions \cite{wang2021review}. Many rely on surrogate safety measures (SSMs) embedded within low-dimensional, piecewise linear frameworks, which fail to capture the nonlinear, high-dimensional interactions characteristic of real-world traffic systems \cite{li2024beyond}. As a result, critical vehicle dynamics—such as tire-road friction, suspension behavior, and superelevation effects—are frequently neglected, undermining the accuracy of safety-critical predictions.
To overcome these limitations, simulation has become a key research tool due to its ability to replicate traffic scenarios under controlled and repeatable conditions \cite{dosovitskiy2017carla,behrisch2011sumo}. Unlike real-world testing, simulation allows for safe and cost-effective evaluation of safety interventions and driver behaviors across diverse traffic conditions \cite{parra2020validation}. For example, Varga et al. \cite{varga2023optimizing} demonstrated how integrating mesoscopic traffic simulation with high-fidelity vehicle dynamics models enhances both computational efficiency and scenario scalability—allowing for extensive testing of critical safety events such as lane changes and emergency braking. This capacity makes simulation particularly valuable for analyzing low-probability, high-severity events that are difficult to observe empirically \cite{wang2018combined}.

Recent advancements in digital twin technology have significantly enhanced simulation capabilities by integrating real-time data and enabling dynamic, high-fidelity representations of traffic systems \cite{ali2022digital,schwarz2022role,liu2020sensor,wang2022mobility}. These technologies facilitate continuous synchronization between virtual and physical environments, improving scenario accuracy and enabling adaptive safety assessments under evolving conditions \cite{almeaibed2021digital}. For example, Wang et al. \cite{wang2022mobility} developed a comprehensive mobility digital twin architecture that supports large-scale traffic management and policy evaluation through real-time data fusion and scenario-based analysis. In another study, Liu et al. \cite{liu2020sensor} introduced a vehicle-to-cloud digital twin framework that integrates connected vehicle data to enhance advanced driver assistance systems (ADAS). Similarly, Wang et al. \cite{wang2021digital} presented a Unity-based digital twin simulation platform capable of analyzing connected and automated vehicle behavior under diverse traffic and environmental conditions. As networks grow more complex, future simulation platforms must integrate heterogeneous data streams, couple physical and behavioral realism, and remain computationally scalable across many scenarios \cite{bhatt2025architecting}.

Based on the challenges outlined, several critical gaps persist in digital twin frameworks for safety analysis. On the real-to-sim side, two main issues emerge. First, there is difficulty in integrating diverse data sources—from high-resolution sensor inputs to detailed static infrastructure data—to construct realistic traffic environments. Second, accurately modeling vehicle dynamics remains challenging, particularly in reproducing realistic driving behavior that accounts for multidimensional factors such as road geometry and direction. On the sim-to-real side, current active safety frameworks often fall short in producing risk assessments within digital twins that reliably reflect the complexities of real-world conditions. These shortcomings highlight the need for systematic approaches to enhance simulation fidelity and improve the reliability of risk evaluations, laying the groundwork for more robust safety solutions.

Our digital-twin platform directly addresses these limitations by integrating high-fidelity road geometry data from aerial LiDAR, multi-sensor vehicle data, realistic traffic flow modeling via Simulation of Urban MObility (SUMO), detailed sensor-level simulation with CAR Learning to Act (CARLA), and advanced vehicle dynamics through NVIDIA PhysX. This system creates a robust, extensible environment for evaluating vehicle behaviors and active safety systems in a variety of realistic, reproducible, and scalable scenarios—paving the way toward safer, smarter, and more inclusive mobility.

\section{Platform Description}
The digital-twin platform integrates four critical components: aerial LiDAR data collection and processing, SUMO traffic flow simulation, CARLA high-fidelity vehicle simulation, and NVIDIA PhysX vehicle dynamics modeling. The aerial LiDAR data collection process involves capturing accurate road geometries, including curvature, cross-section grades, and vertical alignments, which are processed using advanced semantic segmentation and AI classification algorithms. These refined datasets are integrated with OpenStreetMap (OSM) data to create detailed, georeferenced 3D road models within RoadRunner, a simulation preparation tool.

SUMO, responsible for traffic flow management, simulates large-scale network behaviors through realistic routing, lane-changing models, and traffic signal interactions. CARLA complements SUMO by providing a high-resolution, sensor-rich simulation environment crucial for detailed autonomous vehicle analysis. Vehicle dynamics, an essential aspect of realistic simulation, are handled by NVIDIA PhysX, offering precise calculations for suspension, tire-road interactions, and vehicle stability based on varied vehicle configurations.

\begin{figure}[ht]
  \centering
  \includegraphics[width=0.45\textwidth]{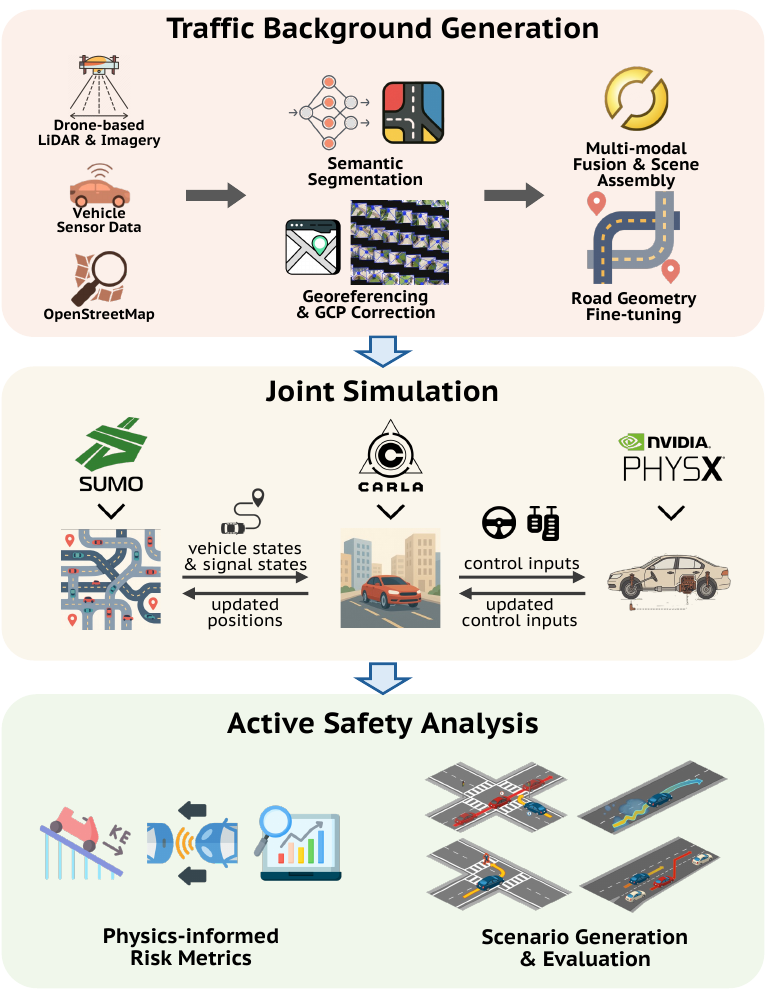}
  \caption{Digital Twin Pipeline.}
  \label{fig:pipeline}
\end{figure}

The joint-simulation architecture synchronizes these components, ensuring consistency between macro-level traffic flow (SUMO) and micro-level vehicle behaviors (CARLA and PhysX), enhancing the realism and reliability of simulations. This robust integration supports comprehensive active safety analysis, enabling detailed scenario testing and informed decision-making. The digital twin pipeline is demonstrated in Fig.~\ref{fig:pipeline}.

\section{Module Description}

\subsection{Traffic Background Generation by Multimodal Data} \label{sec:process}
\begin{figure}[ht]
  \centering
  \includegraphics[width=0.5\textwidth]{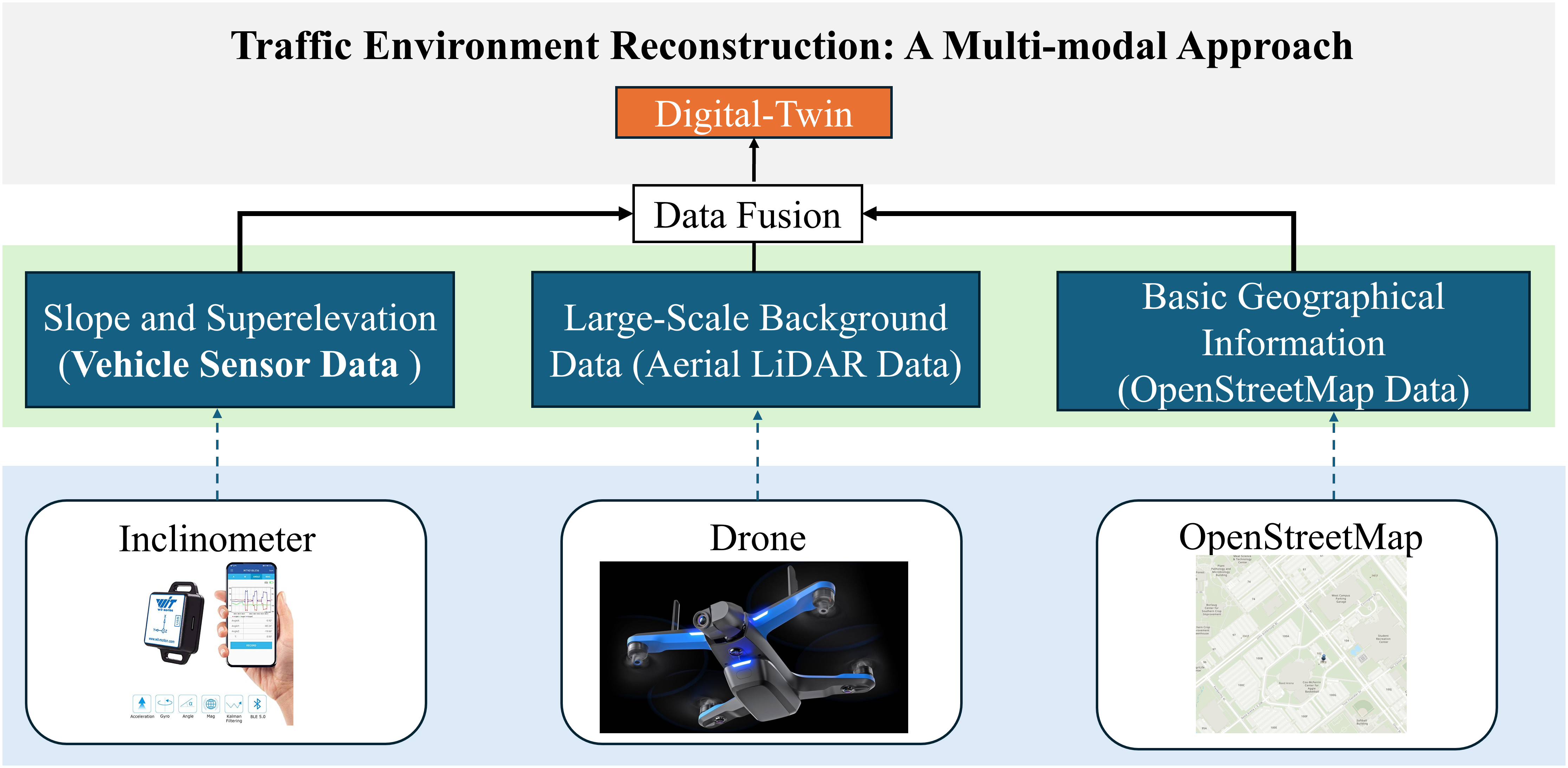}
  \caption{The multi-modal data-enabled traffic environment reconstruction.}
  \label{fig:muti-modal}
\end{figure}

To construct high-fidelity simulation environments, our approach integrates a multi-modal data-enabled traffic environment constructed from drone-based aerial LiDAR, OSM road network data, and vehicle sensor data (e.g., GPS and inclinometer readings), as shown in Fig. \ref{fig:muti-modal}. We used OSM data for the macroscopic generation of the road network structure, capturing features such as intersections and road length. The drone-based aerial LiDAR data was responsible for mesoscopic terrain modeling and the initial construction of the digital twin. Finally, vehicle sensor data was incorporated to perform microscopic refinement of roadway details, including slope and superelevation. The primary equipment includes Unmanned Aerial Vehicles (UAVs) for capturing high-resolution images (i.e. Skydio 2+), RealityCapture software for dense 3D reconstruction, powerful computing hardware, including high-performance CPU and GPU, to process large point clouds and textured meshes, and a multi-sensor device comprises a 10-axis high-stability inclinometer and a high-precision GPS. By fusing these complementary sources, we are able to reconstruct a more accurate and topologically faithful 3D map than would be possible with any single modality alone. This multi-modal fusion enables better alignment between simulated environments and the real world , which is critical for simulating high-fidelity vehicle dynamics, sensor simulation, and safety evaluation.

\subsubsection{Multi-modal Digital Data Collection for Traffic Environment Reconstruction} \label{sec:data}
\begin{figure}[ht!]
  \centering
  \includegraphics[width=0.43\textwidth]{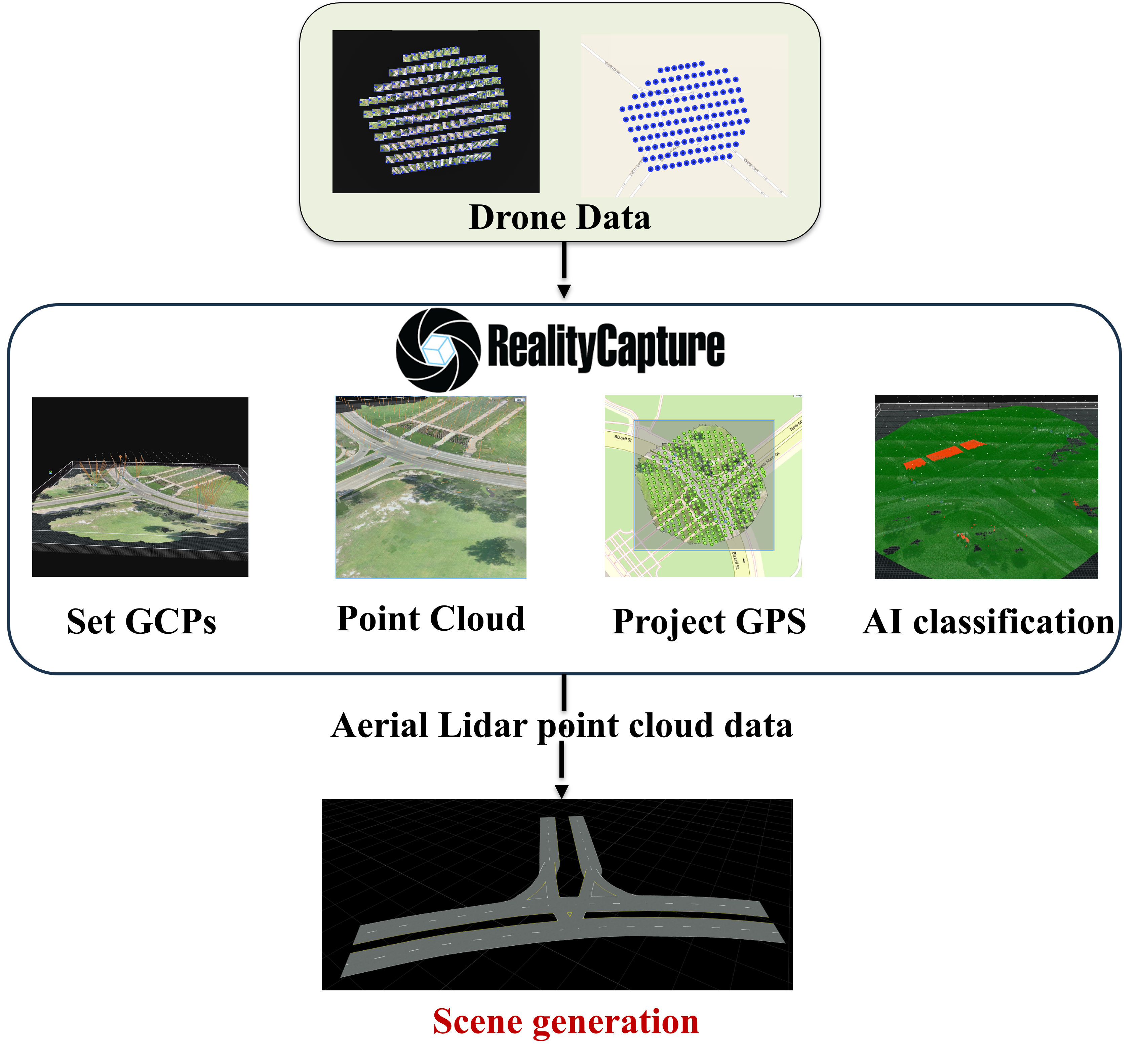}
  \caption{The reconstruction of the drone-scanned data.}
  \label{fig:RC}
\end{figure}

Initially, drone-based LiDAR mapping captures detailed road geometry and environmental context. To ensure georeferenced accuracy and correct potential distortions, such as the bowl effect \citep{jaud2018suggestions}, strategically placed Ground Control Points (GCPs) are used during the RealityCapture-based reconstruction process \citep{barreiro2024validating}. When a real-world road surface that is supposed to be straight is modeled in 3D, distortions can accumulate, resulting in a phenomenon known as the ‘bowl effect’. The model curved to a bowl shape after the alignment. This effect can distort the 3D model, but by incorporating the GCPs, we were able to improve the precision of both the reconstructed 3D model and the LiDAR point cloud data. By marking the same GCP across multiple images, RealityCapture software utilizes different camera angles to triangulate the precise 3D position of each point, ensuring a more accurate reconstruction. The processing workflow for drone-collected data used to generate a LiDAR point cloud and reconstruct a 3D road environment model is illustrated in Fig.~\ref{fig:RC}.

\begin{figure}[ht]
  \centering
  \includegraphics[width=0.3\textwidth]{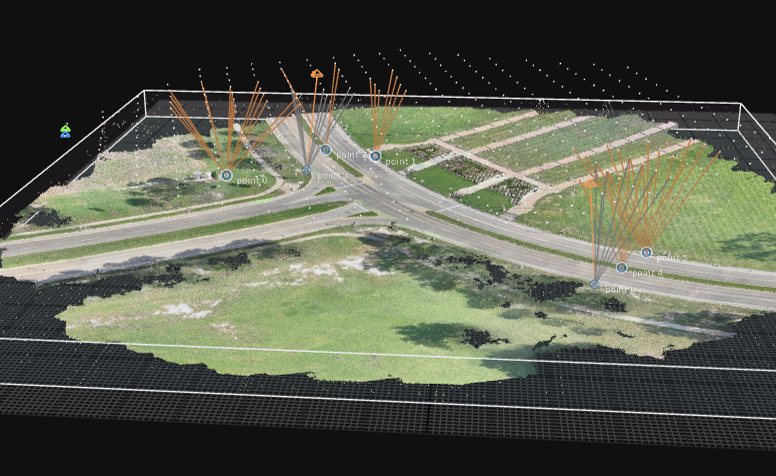}
  \caption{Georeferenced data and GCPs embedded LiDAR point cloud.}
  \label{fig:GCP}
\end{figure}

The drone-collected imagery was processed in RealityCapture to generate a dense 3D point cloud and textured mesh of the environment \citep{realitycapture}. The built-in AI classification function of RealityCapture is employed to classify road surfaces, buildings, and vegetation. Non-road objects are removed to ensure the relevance of the final dataset. Then, an ortho projection is created for the LiDAR point cloud with georeferenced data embedded, which allowed us to embed the georeferenced data for further use and analysis in subsequent project stages. RealityCapture can output 3D model files in various formats, such as \texttt{.las}, \texttt{.obj}, \texttt{shapefile}, and \texttt{AutoCAD DXF}. These files can be imported into CARLA, ArcGIS, and Roadrunner for further design and application. An example is shown in Fig. \ref{fig:GCP} 

Simultaneously, vehicle-mounted multi-sensor systems, comprising high-precision GPS and digital inclinometers, are used to measure vertical alignment grades, cross-slopes, and curvatures of roads. The multi-sensor device comprises a 10-axis high-stability inclinometer and a high-precision GPS, capable of generating time series data including time, acceleration, velocity, angular measurements, magnetism, GPS coordinates, satellite count, and quaternion data. The roadway characteristics were reconstructed by analyzing the attitude and trajectory of the vehicle. The digital multi-sensor device was installed in the trunk of the test car and was placed in such a way that it laid flat or at a neutral level. Once the GPS established a connection with the satellites and all readings of the meters stabilized, roadway data collection began. To achieve precise measurements of the test vehicle’s attitude and trajectories during driving, the test vehicle maintained a speed below 20 mph and stayed in the middle of the lane. The sampling frequency was set to 5 Hz. These measurements can enhance the geometric realism of the RoadRunner-generated scene by fine tuning the map with the measured road geometry data.


\subsubsection{Construction Process of the 3D Digital Twin Map}
\begin{figure}[ht]
  \centering
  \includegraphics[width=0.5\textwidth]{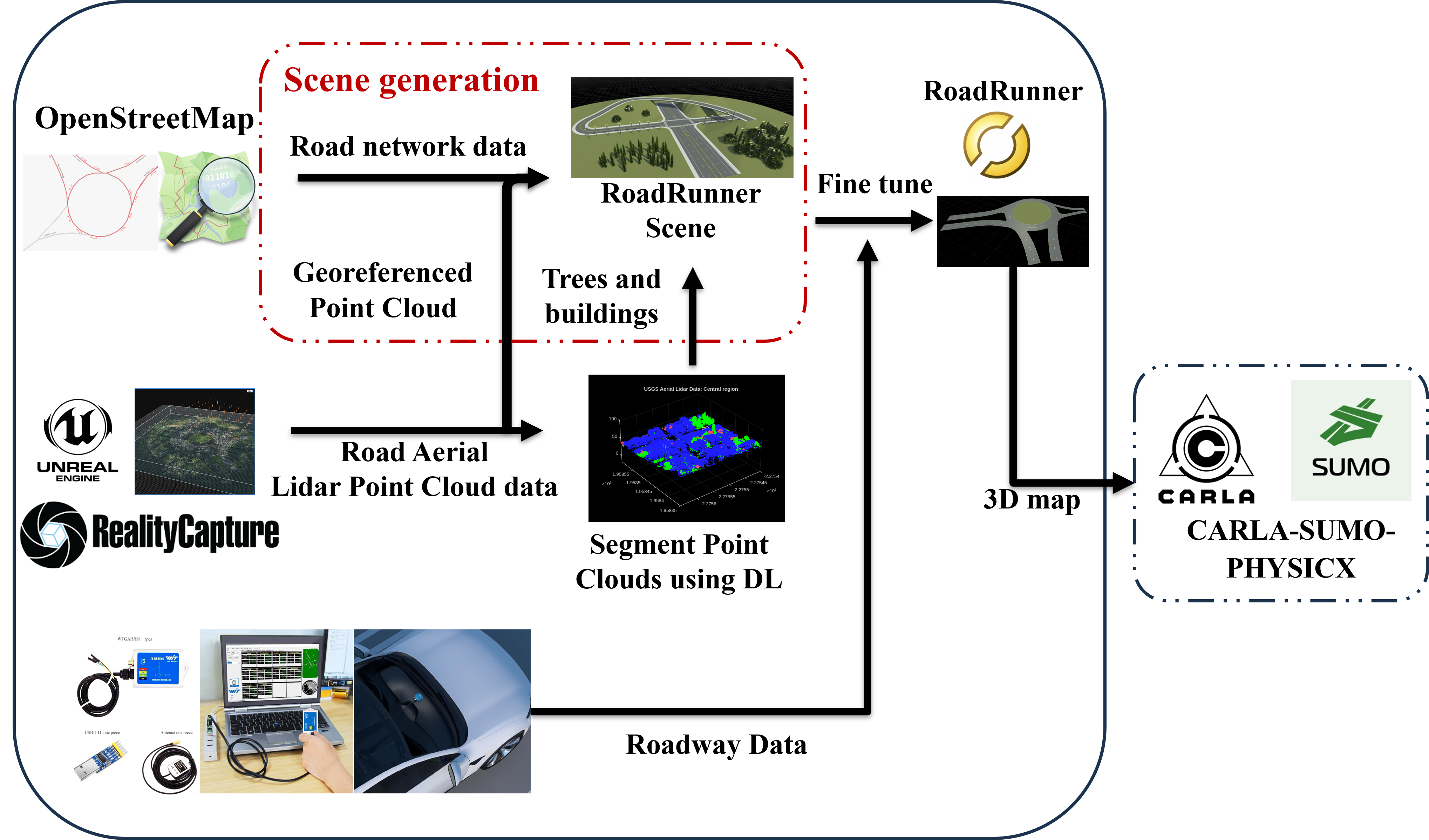}
  \caption{Workflow diagram of the digital infrastructure.}
  \label{fig:generation}
\end{figure}

After gathering the generated 3D point cloud and measuring the roadway data, we can construct the 3D map further by integrating it with additional OSM road network data and importing it into a modeling environment. As shown in Fig. \ref{fig:generation}, our digital twin 3D map construction process consists of three major tasks: a) Generating RoadRunner Scene Using Aerial LiDAR Data and OSM data, b) Customizing and fine tuning the \texttt{.rrhd} map with the measured roadway data, c) Data Extraction and 3D map output for CARLA and SUMO.

\begin{figure}[ht]
  \centering
  \includegraphics[width=0.45\textwidth]{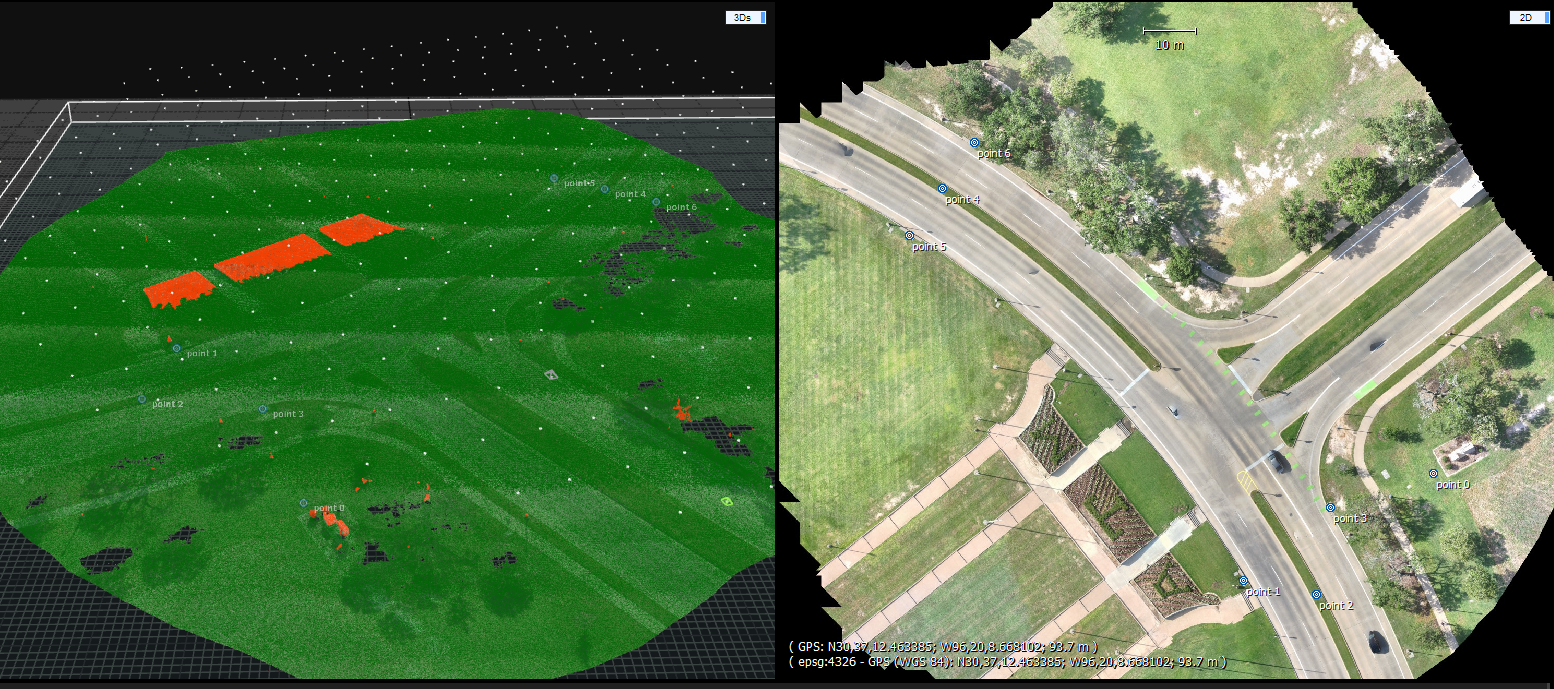}
  \caption{Exemplar semantic segmentation for LiDAR.}
  \label{fig:ai}
\end{figure}

\paragraph{Generating RoadRunner Scene Using Aerial LiDAR Data and OSM data}
To generate the RoadRunner scene, we utilized OSM road network data alongside aerial LiDAR data collected by drones and processed using RealityCapture software. The OSM data provides geo-referenced road network information, including lane widths, the number of lanes, and other key details. The processed aerial LiDAR data, combined with OSM input, are imported into RoadRunner. As shown in Fig.~\ref{fig:ai}, a semantic segmentation network, RandLA-Net \citep{hu2021learning}, is employed to classify objects in the LiDAR data—segmenting roads, trees, buildings, and other scene elements.

\begin{figure}[ht]
  \centering
  \includegraphics[width=0.3\textwidth]{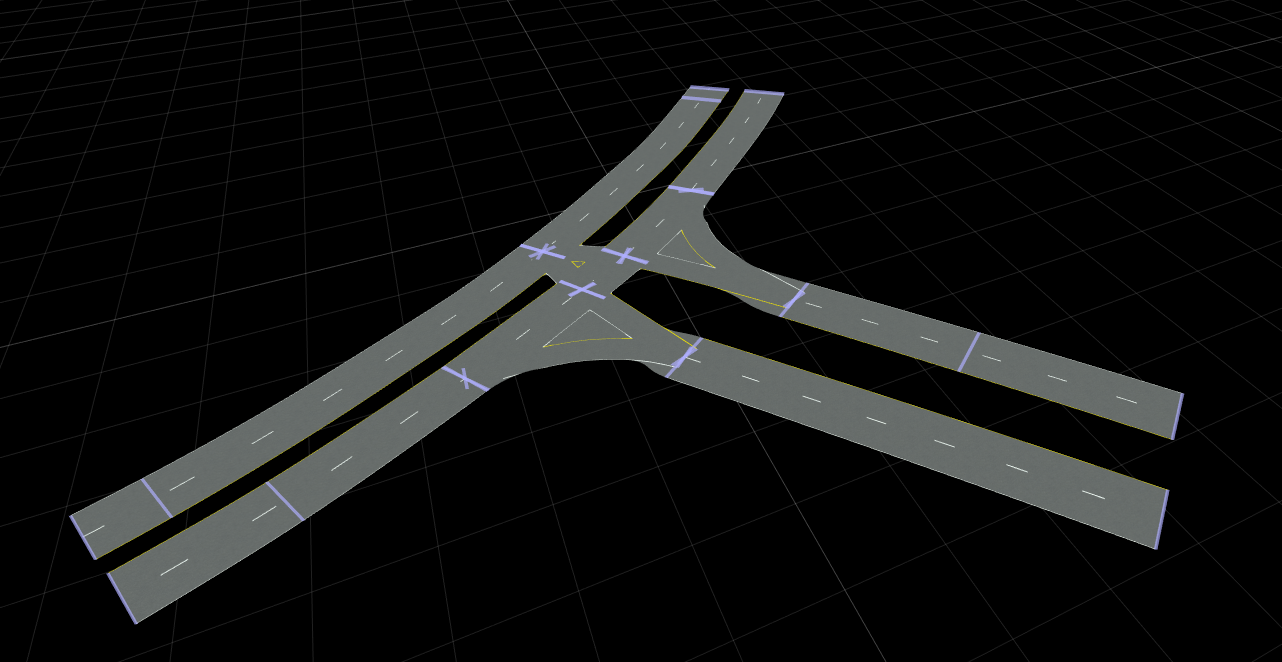}
  \caption{The draft scene of the digital road environment.}
  \label{fig:draft}
\end{figure}

The segmented point cloud is then transformed into a geo-referenced coordinate system, ensuring accurate alignment with the OSM-based road network. Using the RoadRunner Scene Builder tool and the information from the OSM road network, illustrated in Fig. \ref{fig:draft}, we integrated the data processed from the previous stages to generate a sketch of the digital road environment.

Subsequently, detailed geometric attributes—such as lane curvature, slope, and elevation—were extracted and formatted for simulation deployment in both SUMO and CARLA.

\paragraph{Customizing and fine tuning the \texttt{.rrhd} map with the measured roadway data}
The OSM road network data was aligned with the 3D environment to match the general layout and topology of the real-world road network. However, slight discrepancies in geometry were observed between the OSM data and the actual roadway. These errors can stem from the limited positional accuracy of crowd-sourced OSM data, as well as from inaccuracies in the 3D reconstruction process. The accuracy of the 3D model generated by RealityCapture is influenced by several key factors. First, the quality and coverage of the aerial imagery, including lighting conditions, image resolution, and the degree of overlap between images, play a critical role in ensuring reliable feature matching and surface reconstruction. Additionally, the configuration of the RealityCapture software affects the outcome, particularly settings related to alignment quality, camera calibration, and the use of ground control points for georeferencing. Finally, the performance of the processing hardware, such as the capabilities of the CPU and GPU, directly impacts the reconstruction speed and the level of detail achievable in the final 3D model.

In addition to alignment inaccuracies, another challenge arises from incomplete or missing information in the OSM dataset, especially at complex intersections or custom road segments. During the generation of the \texttt{.rrhd} map in RoadRunner, these gaps in OSM data can lead to structural issues in the road network model, such as missing junction points, disconnected lanes, or undefined turning paths. These issues require manual editing and reconstruction in RoadRunner to ensure road continuity and realistic driving behavior.

As a result, the preliminary \texttt{.rrhd} map generated in RoadRunner using aerial LiDAR data and OSM inputs may require fine-tuning. To achieve high-fidelity realism and alignment with the physical world, we adjusted the road geometry by referencing and inputting the roadway data measured by the inclinometer in Section \ref{sec:data}. Also, when an intersection lacks OSM-defined node connections or lane geometry, we need to detect the place of it and inserted the correct junction layout correctly to enable proper behavior at intersections for the subsequent CARLA-SUMO joint-simulation. In our case, we applied a clustering-based approach using density-based spatial clustering of applications with noise (DBSCAN) \citep{ester1996density}. Lane start/end points are projected to 2D and clustered via DBSCAN to identify groups of points corresponding to intersections, whose 3D centroids define junction locations. Lanes are then linked to these junctions based on proximity and directional alignment, and any junction with only a single connection is removed to yield a topologically consistent \texttt{.rrhd} map.

\subsection{Joint-Simulation Platform Construction}

\begin{figure*}[h]
  \centering
  \includegraphics[width=0.8\textwidth]{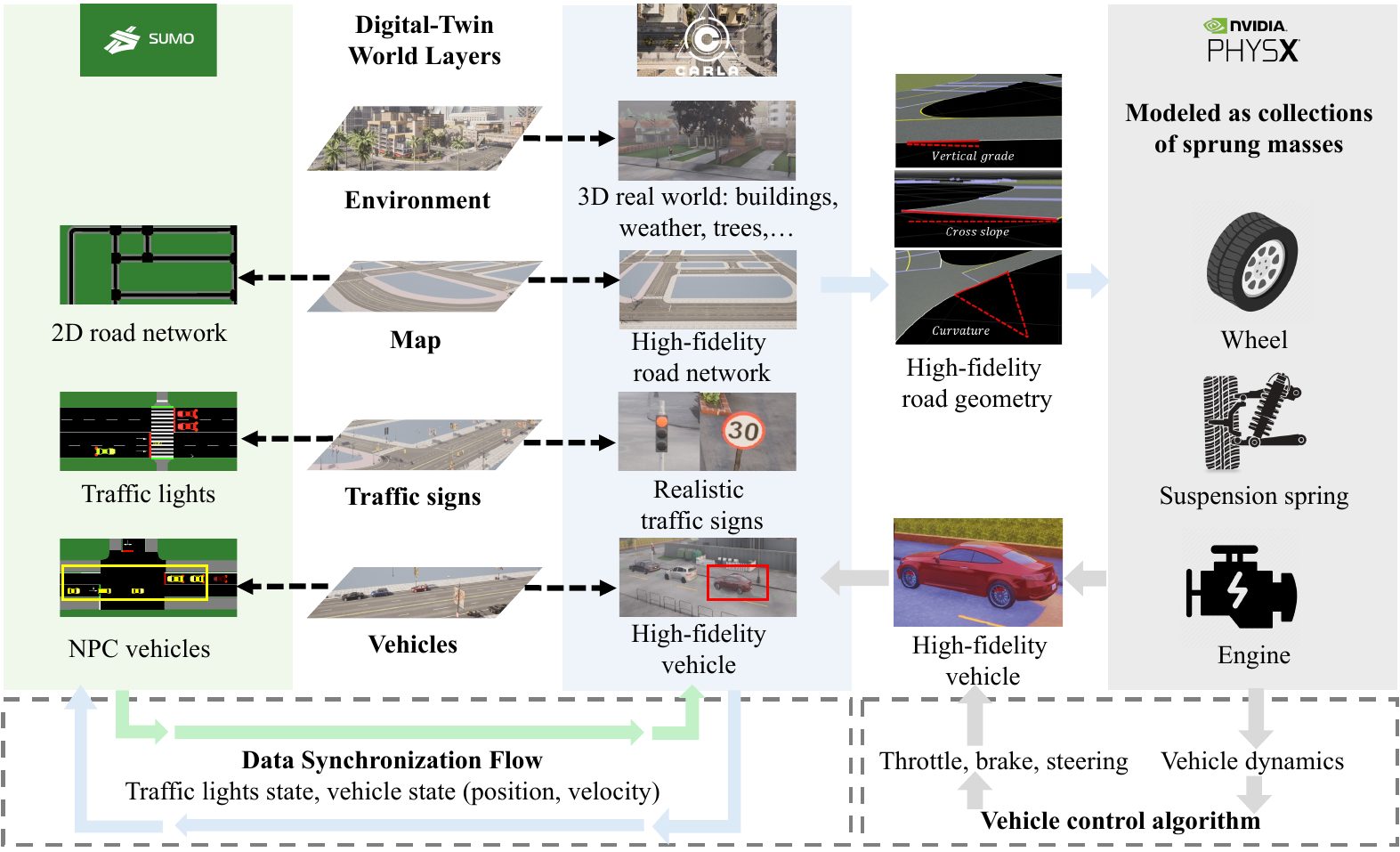} 
  \caption{Joint-simulation platform.}
  \label{fig:co-simulatio-arc}
\end{figure*}

As illustrated in Fig.~\ref{fig:co-simulatio-arc}, the joint-simulation platform integrates CARLA, SUMO, and NVIDIA PhysX. On the left, SUMO represents the 2D road network, traffic signals, and Non-Player Character (NPC) vehicles, enabling large-scale traffic flow simulation. In the center, CARLA simulates a high-fidelity 3D world with realistic roads, traffic signs, and detailed vehicle models—supporting sensor-level realism and advanced perception tasks. On the right, NVIDIA PhysX models vehicle dynamics using a sprung mass system, where wheel, suspension, and engine forces are computed via a raycast-based approach. The dashed rectangles at the bottom of the figure illustrate the data flow between components: SUMO governs NPC traffic and signal states, CARLA manages the sensor-rich 3D environment, and PhysX ensures accurate force calculations for realistic vehicle dynamics simulation. The following subsections describe the primary processes and functionalities of the joint-simulation system.

\subsubsection{Map Reading and Synchronization}

A high-fidelity, georeferenced 3D map is first generated in RoadRunner and saved in \texttt{.rrhd} format. This map is imported into CARLA, which utilizes two exported files: (1) an ASAM OpenDRIVE (\texttt{.xodr}) file, which defines road geometry and topology, and (2) an Unreal Engine Datasmith file (\texttt{.udatasmith}), which contains 3D assets such as buildings and vegetation.

CARLA reads and renders the 3D map. The \texttt{.xodr} file is then converted into a SUMO network file (\texttt{.net.xml}), representing the corresponding 2D road network for use in SUMO simulation.

\subsubsection{Vehicle Spawning and Management}

\paragraph{NPC Vehicle Generation in SUMO}
SUMO spawns NPC vehicles based on user-defined route files and traffic demand specifications, which include vehicle types, departure times, origins, and destinations. Vehicles follow predefined routes or are dynamically assigned via traffic assignment algorithms. Individual vehicle behaviors—including car-following, lane-changing, and interactions with traffic signals—are modeled using microscopic traffic simulation techniques.

\paragraph{Vehicle Generation in CARLA}
CARLA spawns vehicles using detailed blueprints that define appearance, physical attributes, and sensor configurations. It offers a comprehensive suite of onboard sensors such as cameras, LiDAR, RADAR, GNSS, and IMU, facilitating accurate perception, localization, and motion estimation. Vehicle control is implemented through CARLA’s Python API, which allows for direct manipulation of throttle, brake, and steering inputs—enabling the design and evaluation of customized autonomous driving strategies. In this study, a linear feedback control strategy is employed for longitudinal control \cite{zhou2020stabilizing}, while the pure pursuit path tracking algorithm is used for lateral control \cite{samuel2016review}. 

\begin{figure}[ht]
  \centering
  \includegraphics[width=0.43\textwidth]{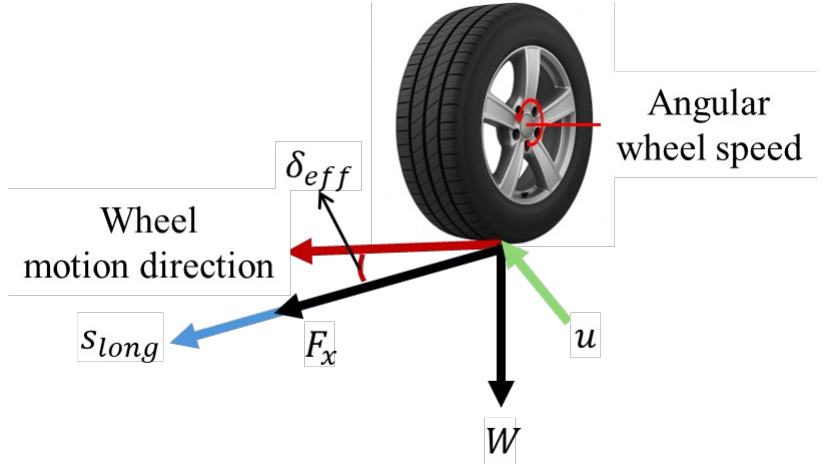}
  \caption{Illustration of terminology in tire models.}
  \label{fig:longitudinal_force}
\end{figure}

\paragraph{Dynamic Simulation Using PhysX}
Once throttle, brake, and steering commands are issued, the PhysX engine processes these inputs to simulate realistic vehicle dynamics \cite{nvidia_physx_sdk_4}. PhysX models vehicles as assemblies of sprung masses, where each suspension line is associated with wheel and tire data. For instance, the longitudinal force \( F_x \) of the tire is computed as:
\begin{equation}
F_x = s_{\text{long}} \times f_{\text{slip}}(s_{\text{combined}}) \times F_0
\label{eq:long_force}
\end{equation}
with the maximum available frictional force given by:
\begin{equation}
F_0 = \frac{\mu \times W}{s_{\text{combined}}}
\label{eq:max_force}
\end{equation}
As shown in Fig.~\ref{fig:longitudinal_force}, \( s_{\text{long}} \) is the longitudinal slip ratio, \( f_{\text{slip}}(\cdot) \) is a smoothing function applied to the combined slip magnitude \( s_{\text{combined}} \), \( \mu \) is the tire-road friction coefficient, \( W \) is the vertical load on the tire. 
The combined slip \( s_{\text{combined}} \) is calculated as:
\begin{equation}
s_{\text{combined}} = \sqrt{s_{\text{long}}^2 + (\zeta \times \tan \delta_{\text{eff}})^2}
\label{eq:combined_slip}
\end{equation}
where \( \zeta \) is a tuning coefficient that scales the lateral slip contribution, \( \delta_{\text{eff}} \) represents the effective slip angle.
This formulation captures the coupling between longitudinal and lateral slip, allowing the model to smoothly saturate the tire force under combined braking/acceleration and cornering conditions. 

The engine drive torque \( T_{\text{engine}} \) is calculated as follows:

\begin{equation}
T_{\text{engine}} = u_{\text{torque}} \times T_{\text{peak}} \times f_{\text{torque}}( \frac{\omega_{\text{engine}}}{\omega_{\text{max}}})
\end{equation}
where $u_{\text{torque}}$ represents throttle input, $T_{\text{peak}}$ is the engine's peak torque,  $f_{\text{torque}}(\cdot)$ is the torque curve function, $\omega_{\text{engine}}$ is the engine's rotational speed, $\omega_{\text{max}}$ is the engine's maximum rotational speed. Please see the source codes of the PhysX for more details of the vehicle dynamics model \cite{nvidia_physx_sdk_4}.


\subsubsection{Synchronization Mechanisms}

\begin{figure}[h]
  \centering
  \includegraphics[width=0.8\columnwidth]{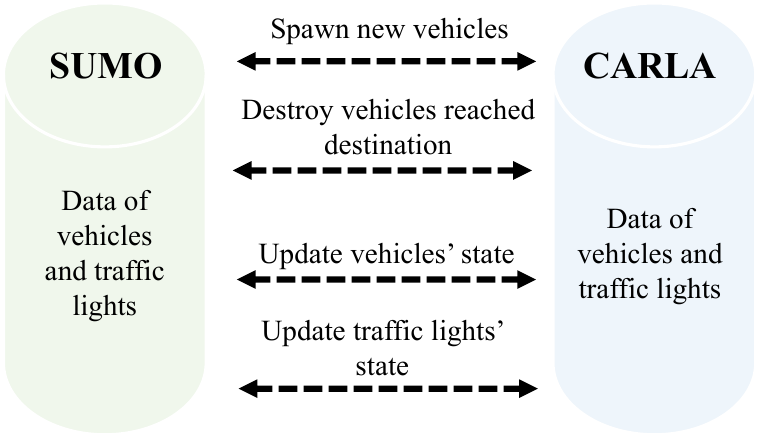}
  \caption{Synchronization between SUMO and CARLA.}
  \label{fig:Synchronization}
\end{figure}

The synchronization between SUMO and CARLA is managed via the \texttt{SimulationSynchronization} class, as depicted in Fig.~\ref{fig:Synchronization}. 

Within CARLA, the \texttt{physics\_control} method integrates the PhysX engine to enhance the realism of vehicle dynamics. The internal methods of \texttt{physics\_control} are detailed in Table \ref{tab:vehicle_physics_control}.

\begin{table}[htbp]
\caption{Methods of \texttt{physics\_control}}
\centering
\renewcommand{\arraystretch}{1.2}
\begin{tabular}{|>{\centering\arraybackslash}m{0.28\linewidth} 
                |>{\centering\arraybackslash}m{0.28\linewidth} 
                |>{\centering\arraybackslash}m{0.28\linewidth}|}
\hline
\textbf{WheelPhysics} & \textbf{VehiclePhysics} & \textbf{GearPhysics} \\
\hline
\makecell{\texttt{tire\_friction} \\ \texttt{damping\_rate} \\  \texttt{radius} \\ \texttt{position}, \ldots} &
\makecell{\texttt{torque\_curve} \\ \texttt{max\_rpm} \\  \texttt{moi} \\ \texttt{mass}, \ldots} &
\makecell{\texttt{ratio} \\ \texttt{down\_ratio} \\ \texttt{up\_ratio}} \\
\hline
\end{tabular}
\label{tab:vehicle_physics_control}
\end{table}

\subsubsection{Joint-Simulation Workflow}

The complete joint-simulation workflow, which sequentially links the core components, proceeds as follows:

\begin{enumerate}
    \item \textbf{Initialize the CARLA simulation:} Launch a high-fidelity 3D environment using the \texttt{CarlaSimulation} class.
    \item \textbf{Convert the CARLA world to SUMO map:} Use the \texttt{netconvert\_carla} function to convert the OpenDRIVE file into a SUMO-compatible network (\texttt{.net.xml}).
    \item \textbf{Initialize the SUMO simulation:} Import the \texttt{.net.xml} file into SUMO using the \texttt{SumoSimulation} class.
    \item \textbf{Spawn vehicles:} Spawn NPC vehicles in SUMO and one or more high-fidelity ego vehicles in CARLA.
    \item \textbf{Start synchronization loop:} Run the simulation in synchronized time steps to maintain consistency between platforms.
\end{enumerate}

\subsection{High-fidelity Active Safety Analysis}
Consider a generic vehicle movement model for the vehicle with index $i$ from the set $V$, given by an ordinary differential equation (ODE):
\begin{equation}
\frac{dZ_i(t)}{dt} = f_i(Z_i(t), u_i(t))
\label{eq:ode}
\end{equation}
where $f(\cdot):\mathbb{R}^n \times \mathbb{R}^m \to \mathbb{R}^n$ is a general state-space function, $Z_i(t) \in \mathbb{R}^n$ represents the state of vehicle $i$ (e.g., position, velocity, heading angle) at time $t$, and $u_i(t) \in \mathbb{R}^m$ denotes the control inputs (e.g., throttle and steering angle) that influence the vehicle's state.

A general framework for multidimensional active safety analysis was proposed by \cite{li2024beyond} to derive the high-fidelity TTC as follows:
\begin{equation}
g_v(Z_i(t_c), Z_j(t_c)) = 0
\label{eq:ttc_1}
\end{equation}
\begin{equation}
g_r(Z_i(t_c), R_k) = 0
\label{eq:ttc_2}
\end{equation}
where $g_v(\cdot)$ characterizes the distance between two vehicles, while $g_r(\cdot)$ characterizes the distance between a vehicle and an obstacle or road boundary. $i \neq j$ and $i, j \in V$, where $V$ denotes the set of vehicles. $R_k \in \mathbb{R}^n$ represents the spatial occupancy of object $k$, and $k \in O$, where $O$ denotes the set of obstacle and boundary indices.
If Eqs.~\eqref{eq:ttc_1} or \eqref{eq:ttc_2} yield non-negative solutions for \( t_c \), a collision is predicted to occur; otherwise, the vehicle is considered free of imminent collision risk. Since TTC is typically evaluated over a rolling time horizon, it is without loss of generality to assume the current time as \( t = 0 \) at which TTC is computed. The control input \( u_i(t) \) depends on specific vehicle or driver behavior models (e.g., adaptive cruise control, car-following models). However, because modeling such behavior is beyond the scope of this study, a constant control input is assumed. When non-negative solutions for \( t_c \) exist, the high-fidelity TTC is defined as the earliest (i.e., minimum) solution and is denoted by \( t_c^{\ast} \).

To illustrate this framework, consider the bicycle model accounting for road slope in a three-dimensional space. The state vector \( Z_i(t) \), and the kinematic equations are given in the following:
\hspace{-1.5em}
\begin{equation}
\frac{d}{dt}
\begin{bmatrix}
x_i(t) \\
y_i(t) \\
z_i(t) \\
\psi_i(t) \\
v_i(t)
\end{bmatrix}
=
\begin{bmatrix}
v_i(t)\cos(\psi_i(t))\cos(\theta(x_i, y_i)) \\
v_i(t)\sin(\psi_i(t))\cos(\theta(x_i, y_i)) \\
v_i(t)\sin(\theta(x_i, y_i)) \\
\frac{v_i(t)}{L} \tan(\delta_i(t)) \\
a_i(t) - g\sin(\theta(x_i, y_i))
\end{bmatrix}
\label{eq:bicycle_slope_3d}
\end{equation}
where \( (x_i, y_i, z_i) \) denote the spatial position of vehicle \( i \), \( \psi_i \) is the heading angle, and \( v_i \) is the longitudinal speed. \( \delta_i \) denotes the steering angle, and \( L \) is the vehicle wheelbase. The function \( \theta(x_i, y_i) \) denotes the inclination (road slope angle) at position \( (x_i, y_i) \), which affects both vertical motion and longitudinal acceleration through gravitational forces. The term \( -g \sin(\theta(x_i, y_i)) \) captures the component of gravity acting along the direction of vehicle motion and the control input \( u_i(t) = [a_i(t),\delta_i(t)]^\top \). This state-space model specifies \( f_i(\cdot) \) in Eq.~\eqref{eq:ode}. To operationalize Eqs.~\eqref{eq:ttc_1} and \eqref{eq:ttc_2}, consider a simple geometric formulation for the function \( g \). Let \( l_i \) and \( l_j \) denote the physical lengths of vehicles \( i \) and \( j \), respectively. Then, the inter-vehicle collision condition can be approximated by a Euclidean distance threshold:
\begin{equation}
g_v(Z_i(t), Z_j(t)) = \|p_i(t) - p_j(t)\|_2 - \frac{l_i + l_j}{2},
\label{eq:gv_example}
\end{equation}
where \( p_i(t) = [x_i(t), y_i(t)]^\top \) is the position of vehicle \( i \) at time \( t \). Similarly, for vehicle–object interactions, let \( R_k \subset \mathbb{R}^2 \) denote the convex bounding region of object \( k \), and define the function \( g_r \) as:
\begin{equation}
g_r(Z_i(t), R_k) = \text{dist}(p_i(t), R_k) - \frac{l_i}{2},
\label{eq:gr_example}
\end{equation}
where \( \text{dist}(p_i(t), R_k) \) denotes the minimum Euclidean distance from the vehicle’s position to the boundary of object \( R_k \). Substituting Eq.~\eqref{eq:bicycle_slope_3d} into Eqs.~\eqref{eq:gv_example} and \eqref{eq:gr_example} allows one to compute \( t_c^{\ast} \) by solving for the earliest collision time—either with another vehicle or an obstacle. The resulting ODEs can be numerically solved using the fourth-order Runge–Kutta method \cite{evans1991new}.


\section{Experiments and Illustrations}
The experimental section comprises four key components that collectively demonstrate the effectiveness of the proposed digital-twin platform. First, we reproduce realistic traffic backgrounds using fused data sources to highlight the importance of multi-modal integration. Second, we showcase examples from the joint simulation platform—integrating CARLA, SUMO, and NVIDIA PhysX—to illustrate the system's ability to capture both micro- and macro-level traffic behaviors. Third, we evaluate active safety analysis components, focusing on vehicle responses under various risk scenarios. Finally, an ablation study is conducted to assess the contribution of each module to the overall system performance.

\subsection{Traffic background reproduction}
As discussed in Section \ref{sec:process}, the traffic background reproduction includes RealityCapture reconstruction for aerial LiDAR point cloud generation, reconstruction RoadRunner scene using aerial LiDAR data and OSM data, fine-tuning the \texttt{.rrhd} map with the measured roadway data, customizing and assigning the missing junction points for the intersection, and data extraction and 3D map output for CARLA and SUMO. Here, we further shows some quantitative results about the traffic background reproduction.





\subsubsection{Fine-Tuning the \texttt{.rrhd} Map Using Measured Roadway Data}
Using the dense 3D point cloud generated from RealityCapture and OSM road network data, we constructed an initial base model of the road network. The OSM road data was aligned with the 3D environment to match the general layout and topology of the real-world road network. However, to improve the accuracy of the 3D model and address issues arising from incomplete or missing information in the OSM dataset—which can lead to structural problems in the road network model, such as missing junction points (Fig. \ref{fig:draft}), or undefined turning paths—customization and fine-tuning processes are employed.

In RoadRunner, a junction represents a spatial region where two or more roads or lanes intersect. Although it does not define lanes directly, it serves as a container or anchor point for connecting them. Intersections can be inconsistently represented without properly assigned junctions, resulting in dangling or unconnected lanes. Assigning junctions ensures that the intersection logic is centralized, consistent, and easier to manage.

\begin{figure}[ht]
  \centering

  \begin{subfigure}[b]{0.48\textwidth}
    \includegraphics[width=\textwidth]{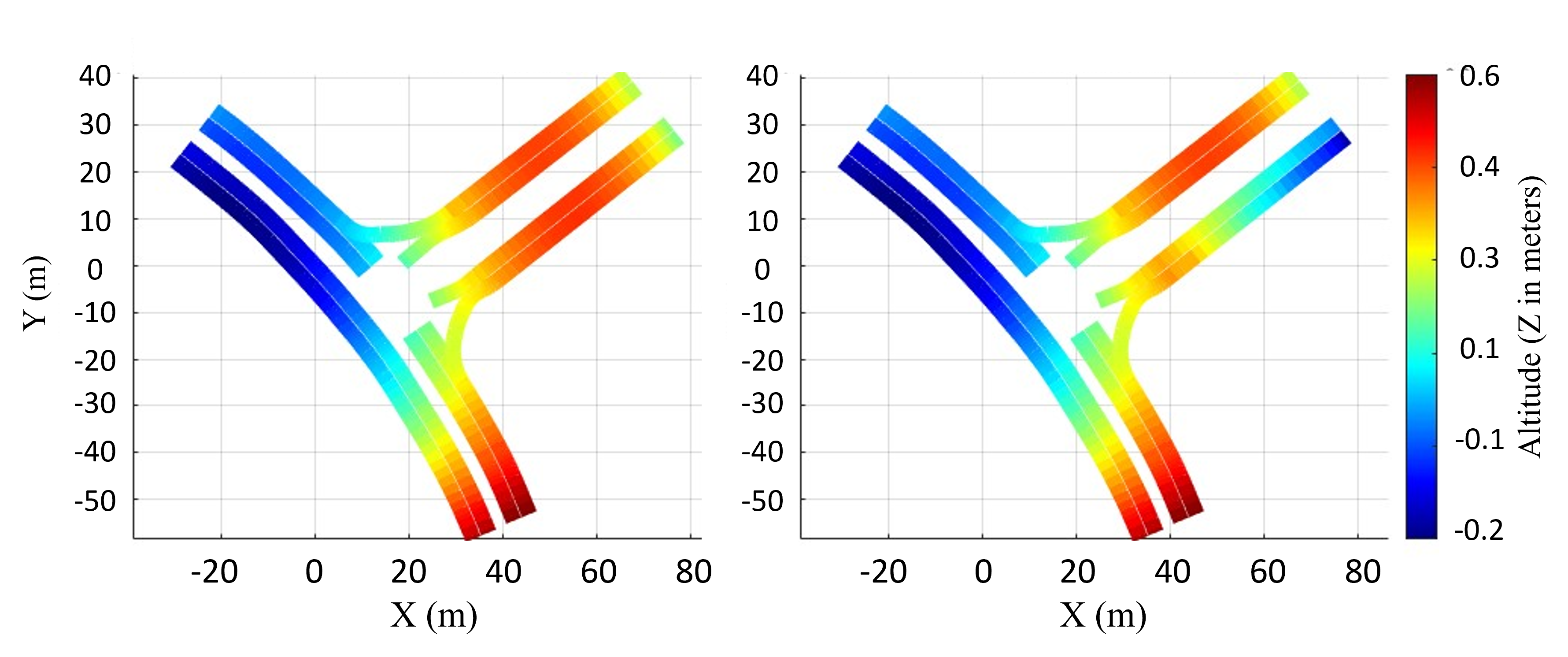}
    \caption{Exemplar roadway terrain visualization (Before: Left, After: Right).}
    \label{fig:fine-tune-a}
  \end{subfigure}
  \hfill
  \begin{subfigure}[b]{0.48\textwidth}
    \includegraphics[width=\textwidth]{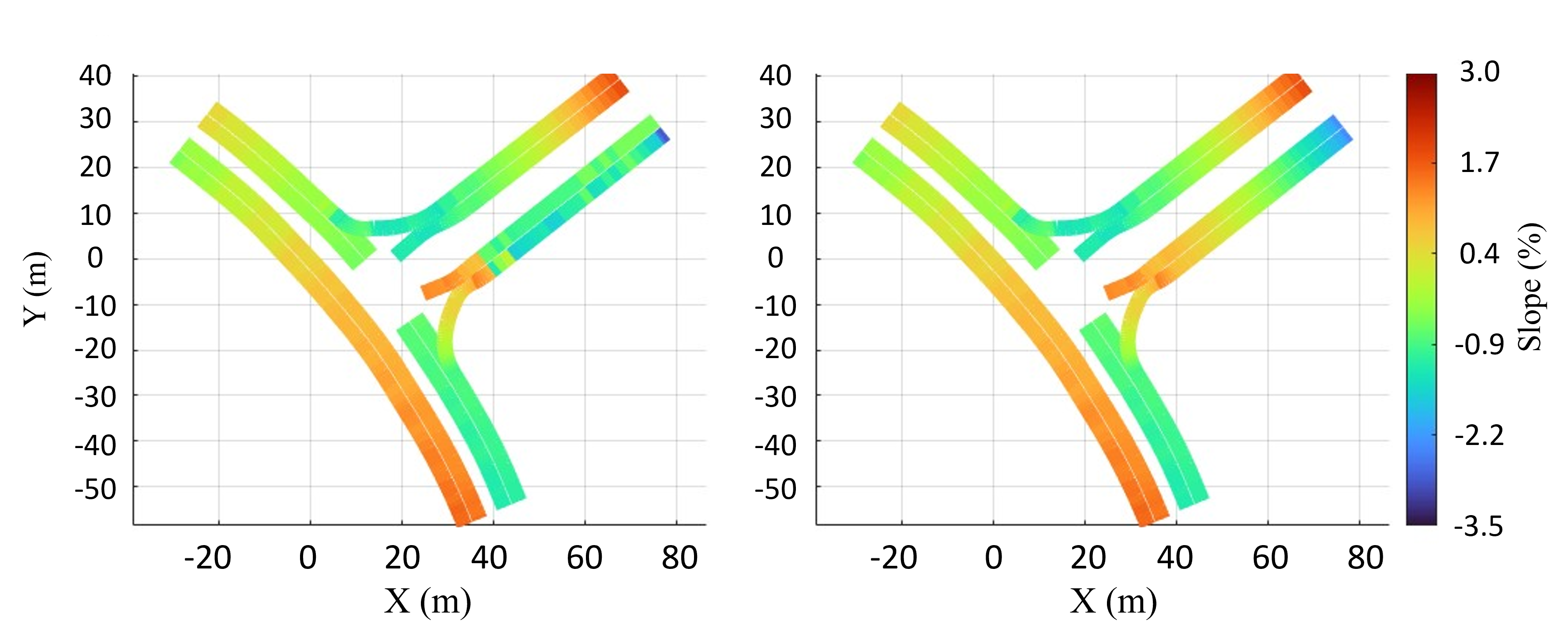}
    \caption{Exemplar roadway slope visualization (Before: Left, After: Right).}
    \label{fig:fine-tune-b}
  \end{subfigure}

  \caption{Roadway terrain adjustment based on inclinometer data fine-tuning.}
  \label{fig:fine-tune}
\end{figure}

Accordingly, the road geometry is first adjusted by referencing and incorporating the inclinometer-measured roadway data. Fig. \ref{fig:fine-tune} relatively represent the roadway terrain and slope visualization. When compared to the original map, minor inaccuracies may appear near the boundaries, but the overall elevation data remains reliably accurate.

\begin{figure}[h]
  \centering
  \includegraphics[width=\columnwidth]{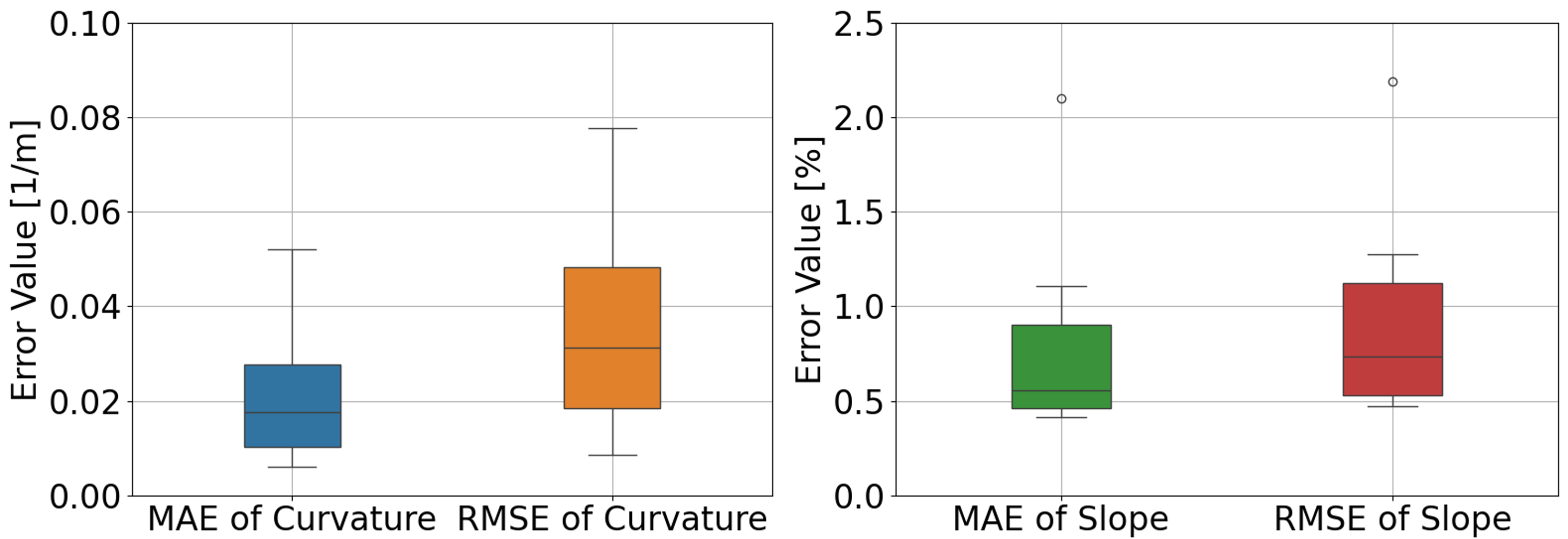}
  \caption{MAE and RMSE for curvature and slope.}
  \label{fig:box1}
\end{figure}

Fig.~\ref{fig:box1} presents box plots of the mean absolute error (MAE) and root mean square error (RMSE) for curvature and slope, based on data integrated from OSM and aerial LiDAR sources. These metrics quantify the accuracy of geometric roadway attributes extracted from digital twin models when compared to a ground truth reference. Curvature estimates remain relatively consistent and low in error, averaging an RMSE of 0.0351. Overall, the averaged RMSE values across all lanes are 0.0351 for curvature, 0.8862 for slope. These results suggest that the fused OSM+LiDAR approach is effective for capturing lane-level geometric features, particularly curvature, though further refinement may be needed for accurately modeling vertical and lateral grade profiles in complex road segments.


  


  


\subsubsection{Customizing and Assigning Missing Junction Points at the Intersection}
\begin{figure}[ht]
  \centering
  \includegraphics[width=0.5\textwidth]{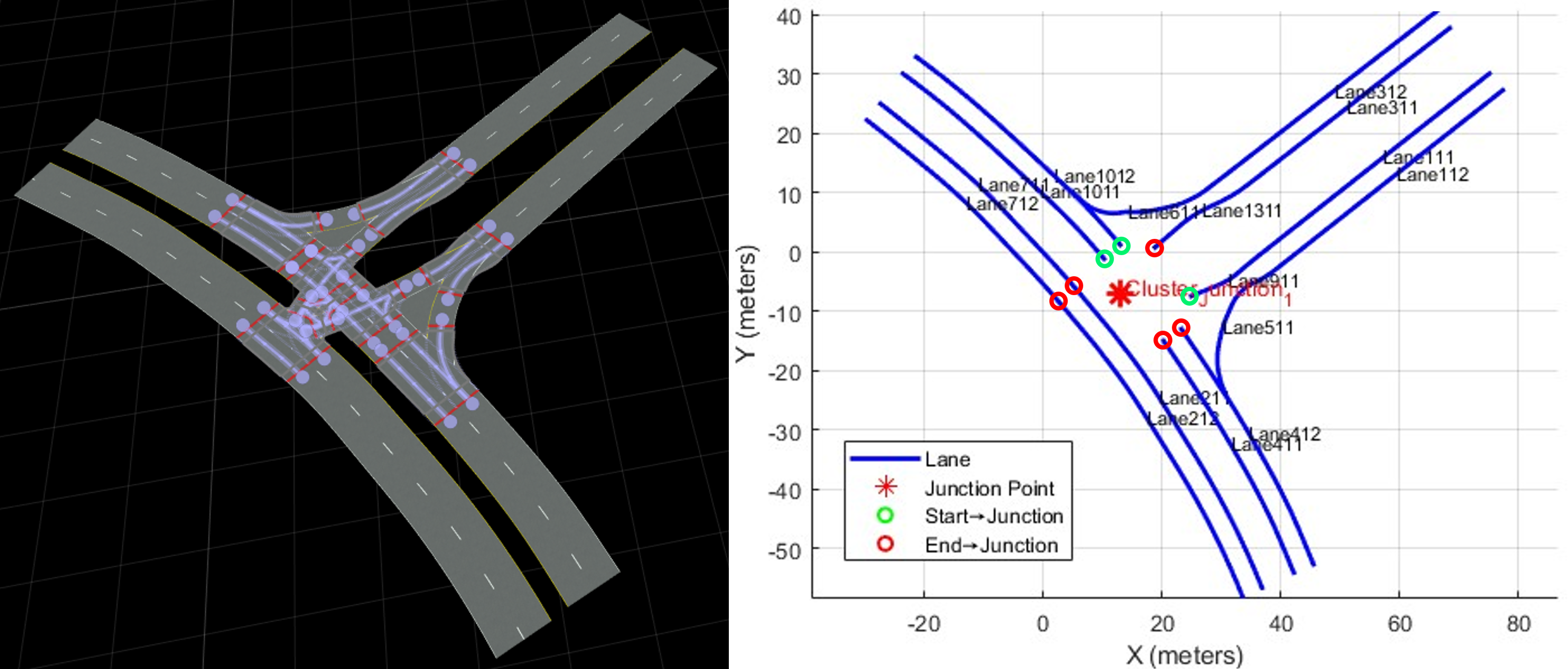}
  \caption{Digital road environment after DBSCAN.}
  \label{fig:junction}
\end{figure}

To address intersections lacking OSM-defined node connections or lane geometry, a clustering-based method using DBSCAN is employed to identify and assign junctions accordingly. Fig. \ref{fig:junction} shows an example of assigned junction point by our Cluster-Based Junction Assignment algorithm. In the graphical user interface (GUI) of Roadrunner (Fig. \ref{fig:junction} left) or \texttt{.rrhd} map (Fig. \ref{fig:junction} right), junctions are often shown as shaded polygons. Assigning them clarifies where complex interactions occur. On the right side of Fig. \ref{fig:junction}, the junction point is shown along with the start and end points of the lanes that connect to it.

\subsubsection{Data Extraction and 3D Map Generation for CARLA and SUMO}
After fine-tuning and validating the \texttt{.rrhd} map, the finalized digital twin environment is prepared for export and integration into both CARLA and SUMO. In addition to simulation use, the geometric data of the roadway can be extracted from the \texttt{.rrhd} file for further analysis, research, or integration into custom pipelines. 

\subsection{Joint Simulation Examples}
\begin{figure}[ht]
  \centering

  \begin{subfigure}[b]{0.49\textwidth}
    \includegraphics[width=\textwidth]{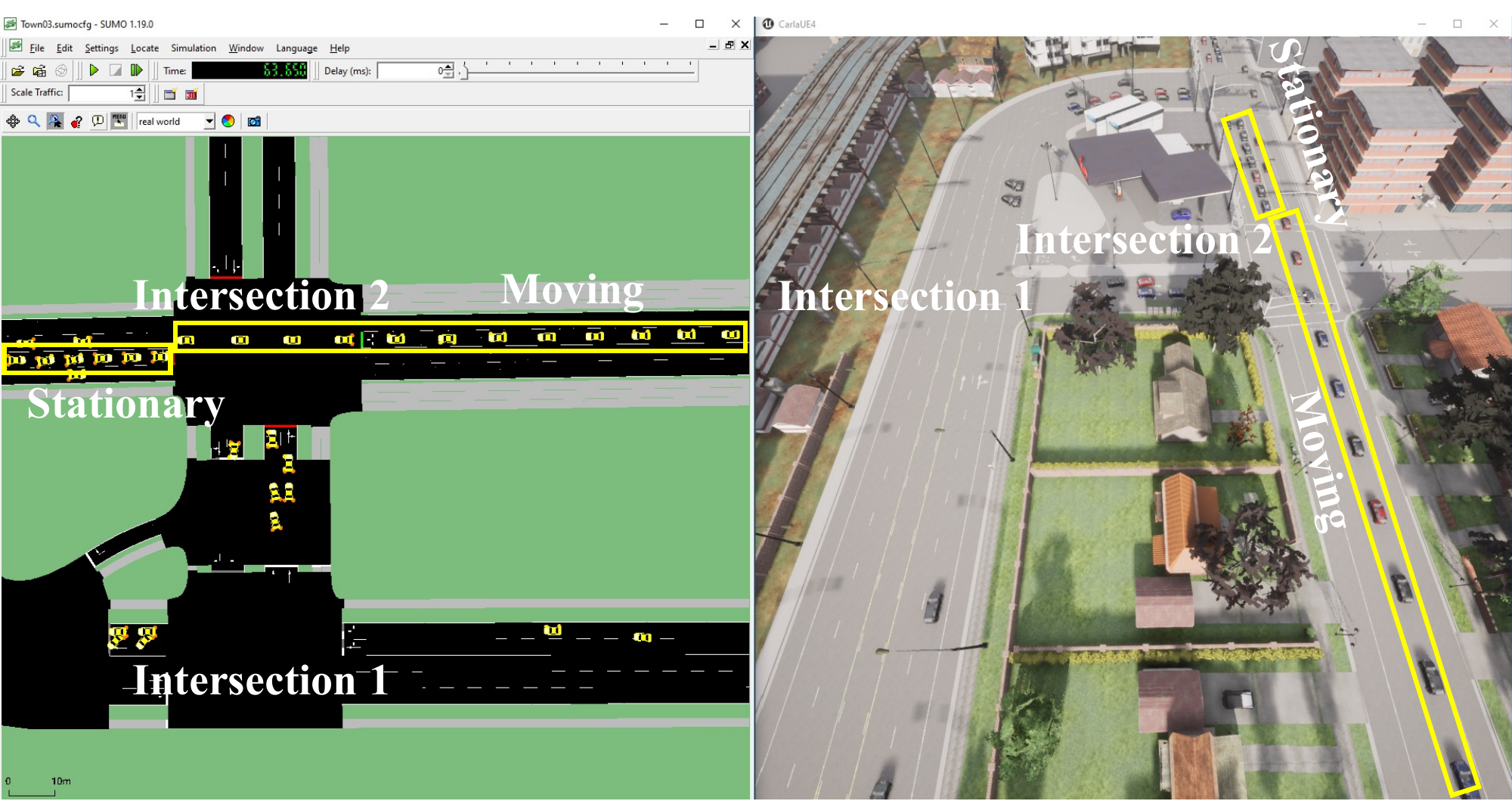}
    \caption{$1\,\mathrm{s}$ desired time gap, $2.5\,\mathrm{m}$ standstill gap}
    \label{fig:sumocarfollowing-small}
  \end{subfigure}
  \hfill
  \begin{subfigure}[b]{0.49\textwidth}
    \includegraphics[width=\textwidth]{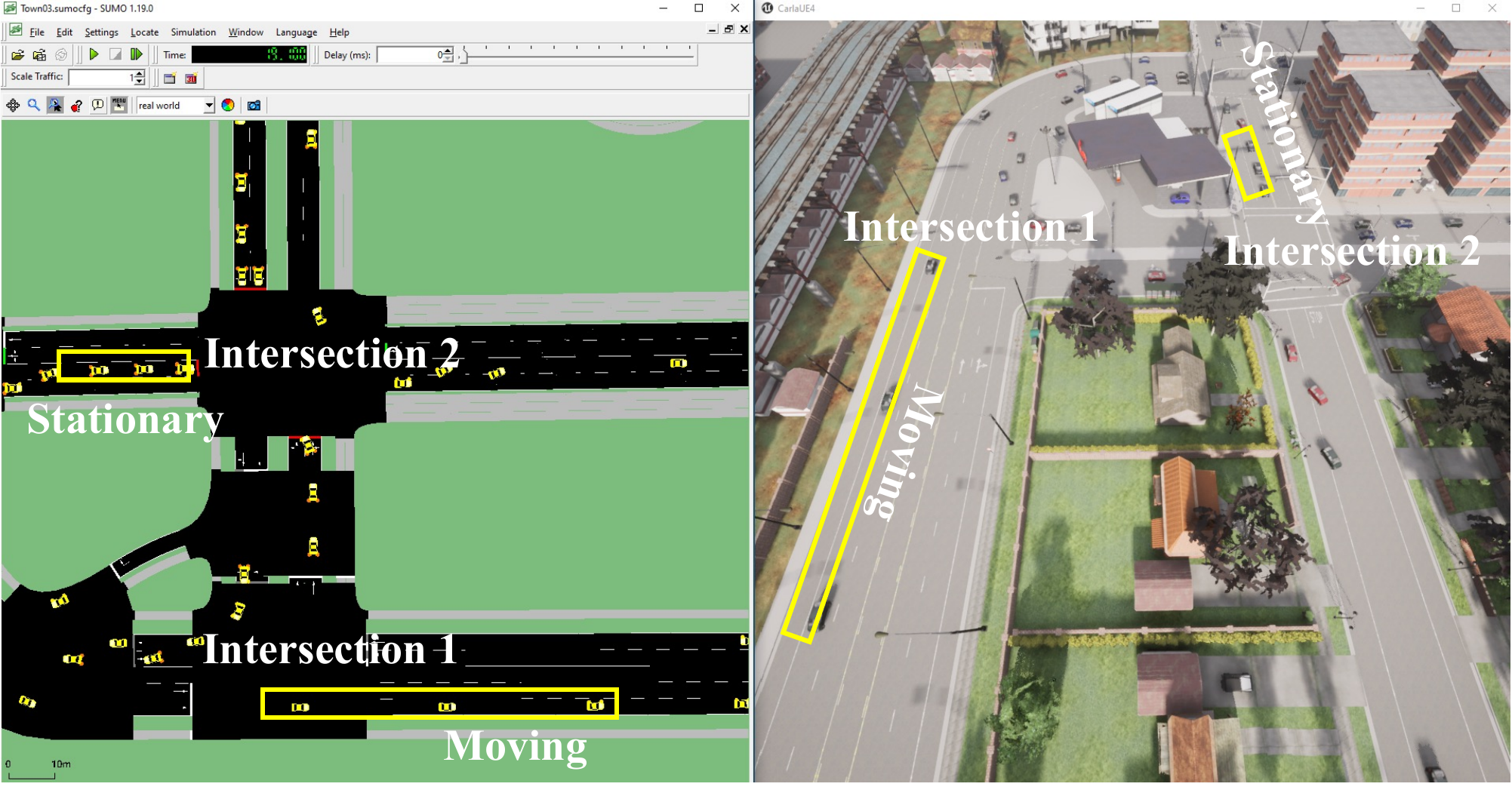}
    \caption{$2\,\mathrm{s}$ desired time gap, $5\,\mathrm{m}$ standstill gap}
    \label{fig:sumocarfollowing-large}
  \end{subfigure}

  \caption{SUMO managing NPC vehicles under different car-following settings.}
  \label{fig:sumocarfollowing}
\end{figure}

Several experiments are conducted to demonstrate the capabilities of the joint simulation platform. Fig.~\ref{fig:sumocarfollowing} illustrates the role of SUMO in managing NPC vehicles within the road network. Specifically, Fig.~\ref{fig:sumocarfollowing} (a) presents vehicles operating under a car-following model with a desired time gap of 1 second and a standstill gap of 2.5 meters (8.2 ft), while Fig.~\ref{fig:sumocarfollowing} (b) shows a longer time gap of 2.5 seconds and a standstill gap of 5.0 meters (16.4 ft). In both subfigures, the yellow lines highlight both moving and stationary car-following platoons. 

\begin{figure}[ht]
  \centering
  \includegraphics[width=0.48\textwidth]{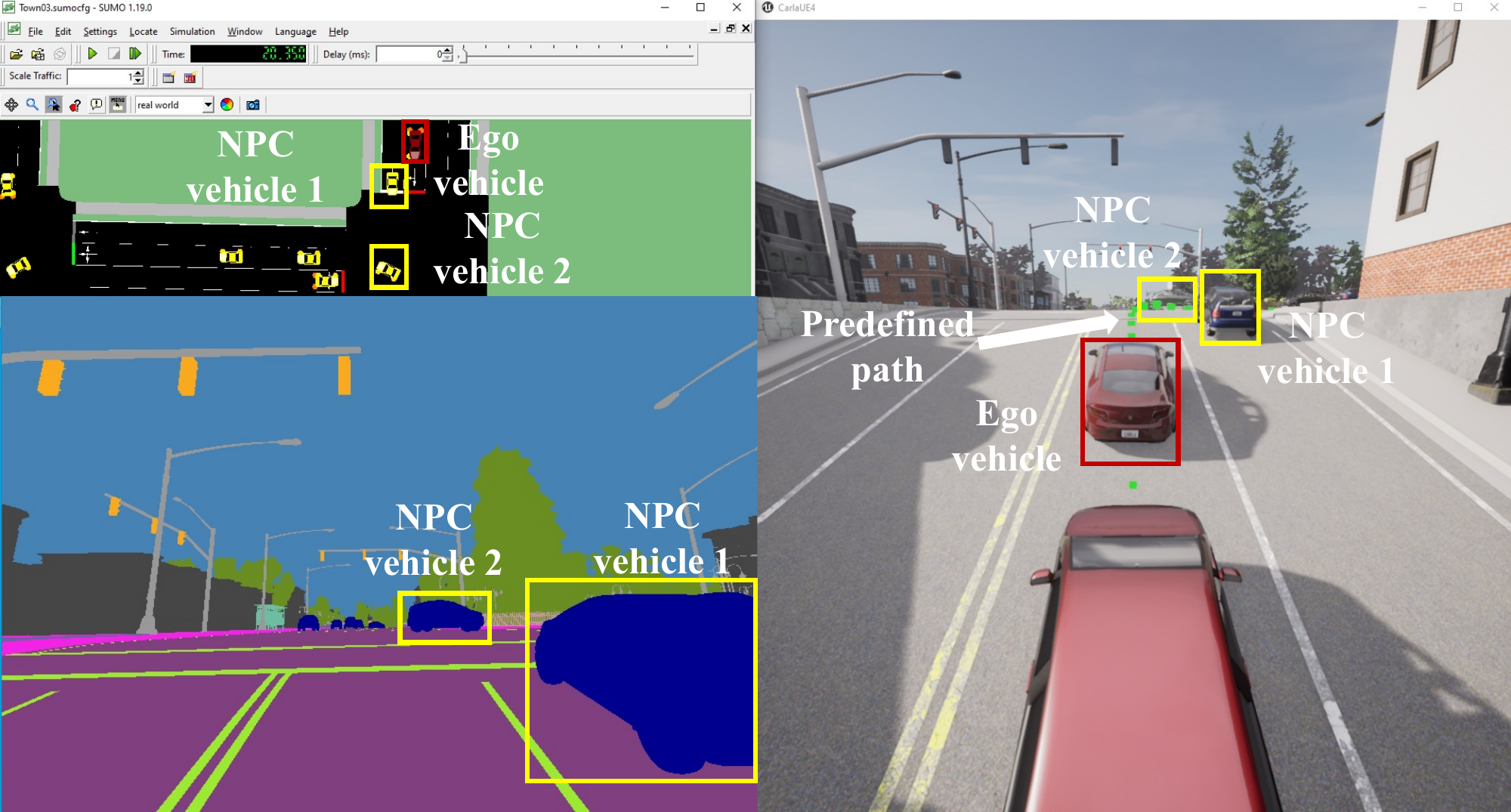}
  \caption{CARLA simulating a realistic environment and ego vehicle.}
  \label{fig:segmentation}
\end{figure}

\begin{figure}[ht]
  \centering
  \includegraphics[width=0.45\textwidth]{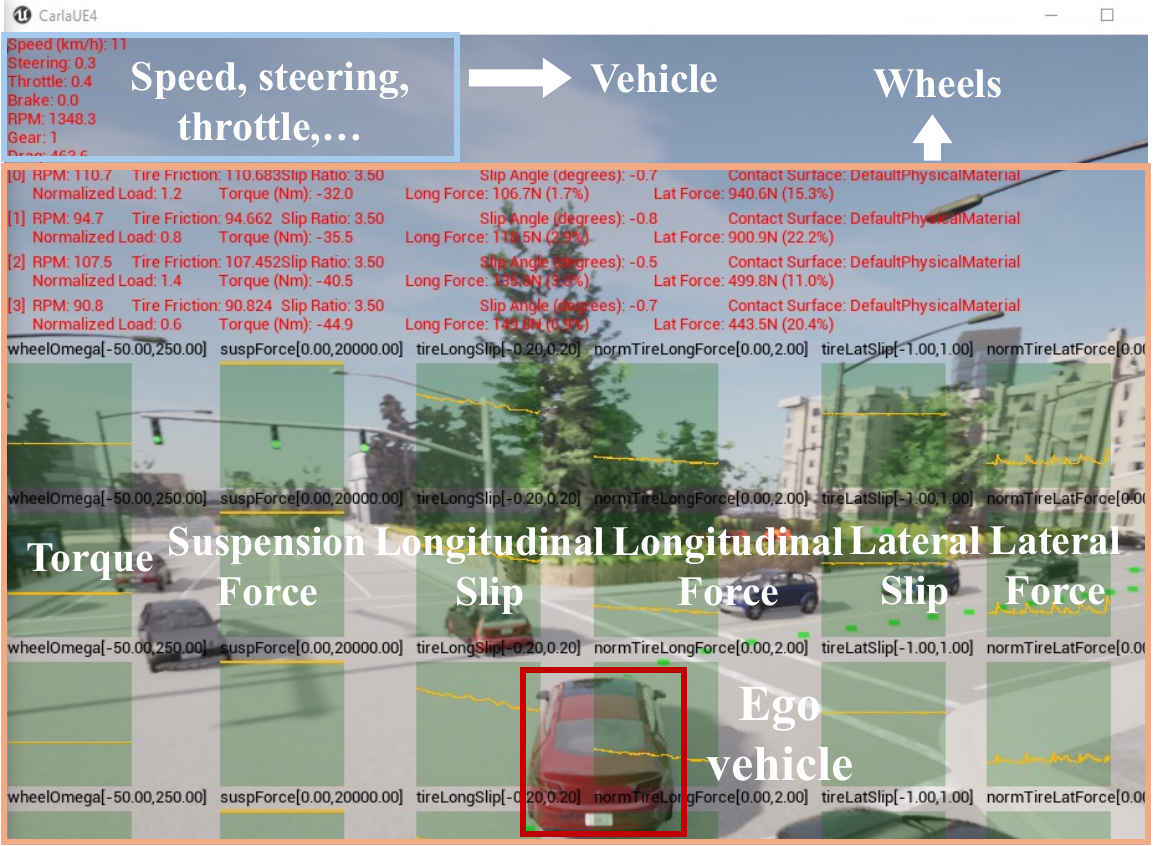}
  \caption{PhysX simulating vehicle dynamics.}
  \label{fig:physicsx}
\end{figure}

Fig.~\ref{fig:segmentation} highlights the role of CARLA in simulating a realistic 3D environment, including high-fidelity maps and the ego vehicle. The sensor view window displays output from a semantic segmentation camera, which classifies each visible object by assigning it a distinct color. Green points indicate the predefined path of the ego vehicle, which is also highlighted in red within the SUMO environment. NPC vehicles detected by the sensors are outlined in yellow. 
CARLA integrates the PhysX engine to simulate vehicle dynamics with high fidelity. Fig.~\ref{fig:physicsx} further presents detailed information provided by the PhysX simulation, such as steering angles, vehicle speed, and the forces and slip angles acting on all four wheels.

\subsection{Active Safety Analysis and Comparison}
To further demonstrate the fidelity of the active safety analysis tool, we employed digital twin techniques to generate eight synthetic collision scenarios spanning various categories, including rear-end collisions, sideswipe collisions, and blind spot incidents. These scenarios are used to compare the traditional TTC metric with the proposed active safety analysis method described in Section~III.

\begin{figure}[ht]
  \centering

  \begin{subfigure}[b]{0.48\textwidth}
    \includegraphics[width=\textwidth]{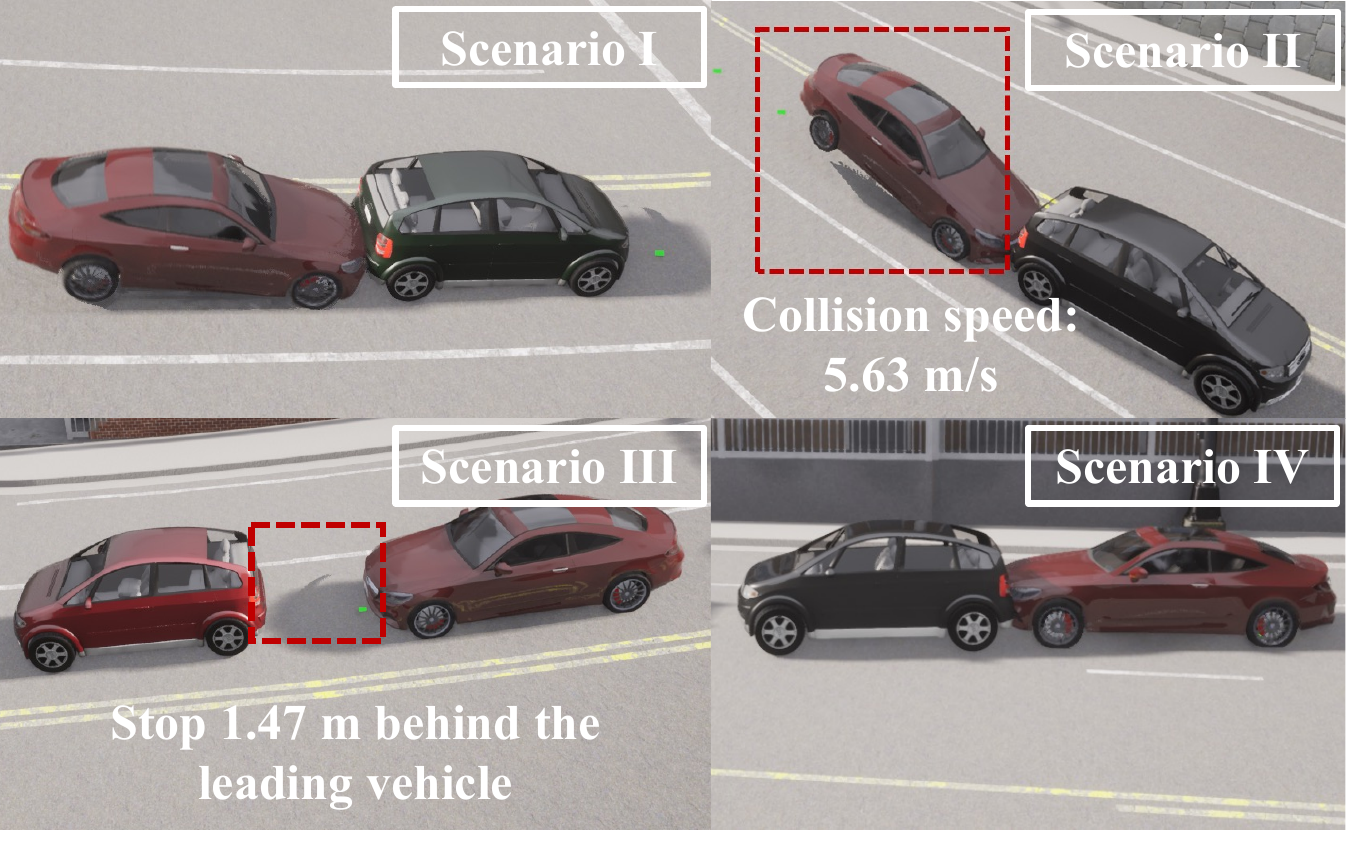}
    \caption{Scenarios I-IV}
    \label{fig:scenario1t4}
  \end{subfigure}
  \hfill
  \begin{subfigure}[b]{0.43\textwidth}
    \includegraphics[width=\textwidth]{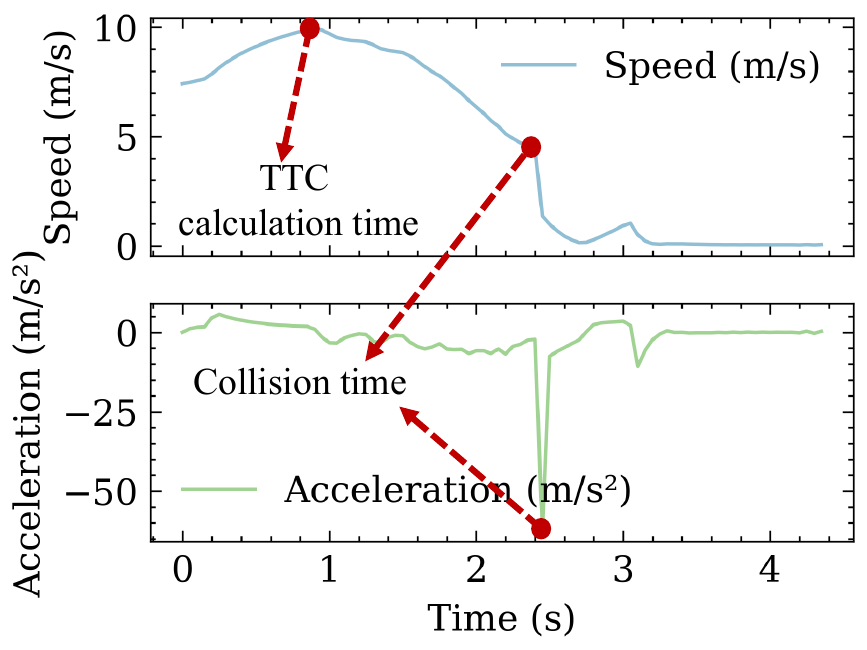}
    \caption{Speed and acceleration profiles of Scenario I}
    \label{fig:scenario1b}
  \end{subfigure}

  \caption{Scenarios I–IV: Rear-end collisions.}
  \label{fig:scenario1}
\end{figure}

Fig.~\ref{fig:scenario1} presents \textit{Scenario I-IV}. \textit{Scenario I} is a rear-end collision on a downhill road with a grade of 0.2. As shown in Fig.~\ref{fig:scenario1} (b), an abnormal deceleration is observed due to the impact force during the collision, indicating the time of collision occurrence. In rear-end scenarios, TTC is evaluated at the moment when the ego vehicle begins to decelerate in response to the lead vehicle. For \textit{Scenario I}, the simulated TTC is 1.06 seconds, while the traditional TTC is 1.60 seconds. This discrepancy arises from the traditional TTC’s assumption of constant speed, which fails to account for road slope and high-fidelity vehicle dynamics. To compute the proposed high-fidelity TTC using the bicycle model, the acceleration \( a_i(t) \) and steering angle \( \delta_i(t) \) are assumed constant and set to their values at \( t = 0 \), the time of TTC evaluation. This approach yields a high-fidelity TTC of 1.35 seconds—closer to the simulated result but still limited by the simplified nature of the bicycle model.

Building upon \textit{Scenario I}, two additional cases are designed to highlight the importance of modeling vehicle dynamics in active safety analysis. \textit{Scenario II} examines the effect of road surface conditions by simulating a low-friction environment (e.g., rain or snow), achieved by reducing the friction coefficient from 5 to 2. For details on the definition of the friction coefficient, refer to the official CARLA documentation \cite{dosovitskiy2017carla}. In this scenario, the simulated TTC decreases to 1.34 seconds, and a more severe collision is observed, with a collision speed of 5.63~m/s compared to 4.81~m/s in \textit{Scenario I}. In \textit{Scenario III}, the vehicle mass is reduced from 1630~kg to 1200~kg, while the initial speed and braking input remain unchanged. As a result, the vehicle successfully avoids a collision and stops 1.47~m before reaching the lead vehicle. \textit{Scenario IV} considers a rear-end collision on a flat road (i.e., zero slope). In all scenarios, the NPC vehicle shares the same model, differing only in randomly assigned colors.

\begin{figure}[ht]
  \centering
  \includegraphics[width=0.41\textwidth]{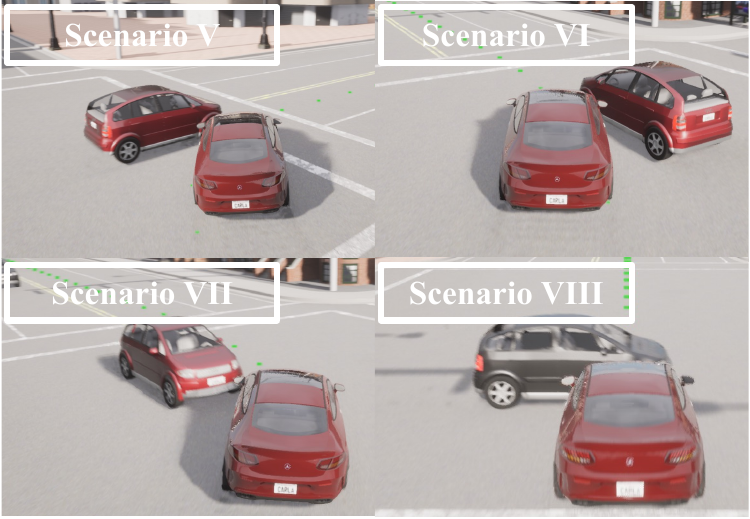}
  \caption{Scenarios V-VIII: Intersection collisions.}
  \label{fig:scenario5t8}
\end{figure}

Intersection-related collisions are analyzed in \textit{Scenarios V}--\textit{VIII}, as shown in Fig.~\ref{fig:scenario5t8}. \textit{Scenarios V} and \textit{VI} simulate sideswipe collisions in the same direction, which typically result from improper lane changes, blind spot misjudgment, or failure to yield during merging. In all intersection scenarios, vehicle speeds are set to remain constant prior to collision to simplify short-term dynamics and represent situations where drivers either fail to perceive the threat or lack sufficient time to react. TTC is evaluated at the time the ego vehicle enters the intersection. \textit{Scenario VII} depicts a Left-Turn Across Path -- Opposite Direction collision, which occurs when a left-turning vehicle crosses the path of an oncoming vehicle traveling straight from the opposite direction. These collisions often stem from misjudged gaps, limited visibility, or inadequate signal control. Finally, \textit{Scenario VIII} illustrates a Right-Angle (Straight-Through) collision, typically caused when a vehicle runs a red light or stop sign and strikes a vehicle traveling perpendicular to its path.

Since \textit{Scenarios II} and \textit{III} involve variations in tire friction and vehicle mass for illustration purposes, their TTC results are excluded from direct comparison with the other scenarios. Table~\ref{tab:ttc_comparison} presents the TTC values and the corresponding difference between calculated and simulated results—for the remaining scenarios. 
\begin{table}[htbp]
\caption{Comparison of Traditional, Simulated, and High-Fidelity TTC Across Scenarios}
\centering
\renewcommand{\arraystretch}{1.2}
\begin{tabular}{|>{\centering\arraybackslash}m{1.5cm} 
                |>{\centering\arraybackslash}m{1.5cm} 
                |>{\centering\arraybackslash}m{1.5cm} 
                |>{\centering\arraybackslash}m{1.5cm}|}
\hline
\textbf{Scenario} & \textbf{Traditional (s)} & \textbf{Simulated (s)} & \textbf{High-Fidelity (s)} \\
\hline
I     & 1.06 (-0.54) & 1.60 & 1.35 (-0.25) \\
IV    & 0.81 (-0.29) & 1.10 & 0.89 (-0.21) \\
V     & 1.70 (+0.40) & 1.30 & 1.12 (-0.18) \\
VI    & 4.12 (+2.17) & 1.95 & 1.35 (-0.60) \\
VII   & 1.27 (-0.33) & 1.60 & 1.86 (+0.26) \\
VIII  & 1.58 (-0.52) & 2.10 & 1.87 (-0.23) \\
\hline
\textbf{Mean Error} & +0.15 & -- & $-0.20$ \\
\textbf{MAE}        & 0.71  & -- & 0.29 \\
\textbf{RMSE}       & 0.97  & -- & 0.32 \\
\hline
\end{tabular}
\label{tab:ttc_comparison}
\end{table}
The high-fidelity TTC consistently achieves lower error magnitudes across all cases. The traditional TTC exhibits high variability, with maximum error reaching 2.17 seconds, compared to a maximum error of only 0.6 seconds for the high-fidelity TTC. On average, the traditional TTC underestimates collision risk, with a mean error of 0.15 seconds, while the high-fidelity TTC tends to slightly overestimate risk, with a mean error of -0.20 seconds. This distinction is safety-critical, as underestimation of collision risk can lead to delayed responses and even crashes. Furthermore, the high-fidelity TTC yields a lower MAE of 0.29 seconds and RMSE of 0.32 seconds, outperforming the traditional TTC, which reports an MAE of 0.71 seconds and RMSE of 0.97 seconds, suggesting better accuracy.

\section{Conclusion}
This paper presents a framework outlining a holistic pipeline for a traffic safety analysis digital twin. The pipeline encompasses traffic background generation, a joint simulation platform for background traffic, AV control customization, high-fidelity vehicle dynamics reproduction, and active safety analysis. Specifically, the framework proposes a multi-modal approach to background generation that combines the easy accessibility of OSM, the detailed data captured by drones, and road slope and superelevation information from vehicle-mounted sensors. The synthesized background is then embedded within a joint simulation platform that integrates SUMO, CARLA, and NVIDIA PhysX. Within this composite system, SUMO generates background NPC vehicles (primarily human-driven), CARLA simulates target AVs or human-driven vehicles with customized algorithms and sensor configurations, and NVIDIA PhysX reproduces vehicle dynamics with high fidelity. In addition, the simulation platform incorporates high-fidelity surrogate safety measures to accurately estimate potential traffic risks. The proposed framework is demonstrated through real-world case studies, showing its capability to reproduce realistic traffic backgrounds. In our examples, the combination of OSM and drone data effectively facilitates background generation. Overall, curvature shows the highest consistency and lowest error across all lanes, indicating strong reliability in horizontal alignment estimation. Slope accuracy is generally acceptable, though with moderate variation depending on the terrain. These findings highlight the effectiveness of the OSM+LiDAR approach for capturing horizontal geometry while some boundary distortions occur, indicating that improvements are still needed for vertical and lateral grade accuracy. These issues are mitigated by vehicle-mounted sensor data.

We further present an illustrative case study based on joint simulation, in which eight synthetic collision scenarios are generated to evaluate our high-fidelity analysis tool against traditional TTC metrics. The results clearly demonstrate that our method significantly enhances the detection and evaluation of potential traffic risks across all scenarios, outperforming traditional TTC by 59\% in mean absolute error and 67\% in root mean square error. 

While the presented framework demonstrates significant advancements in simulating and analyzing traffic safety, several limitations remain. Current models still rely on certain assumptions that may not capture the full complexity of real-world vehicular interactions. Future research could embed advanced AI models to predict intricate vehicle interactions, further enhancing the realism of the simulation \cite{wu2023graph,wu2024hypergraph,zhang2022temporal}.  Another promising direction involves integrating advanced vehicle control algorithms—ranging from classical model-based controllers to reinforcement learning-based approaches—into the simulation to evaluate their effectiveness in maintaining platoon stability and safety under dynamic traffic conditions \cite{zhou2020stabilizing,shi2023deep}. Additionally, leveraging generative AI to produce a wider variety of scenarios—including rare and extreme events—offers a promising direction for diversifying training data and improving overall system robustness \cite{pronovost2023scenario}. These enhancements have the potential to further bridge the gap between simulation and real-world performance in autonomous driving safety analysis.


%

\appendices

\section*{Acknowledgment}
This research is funded by Federal Highway Administration (FHWA) Exploratory Advanced Research 693JJ323C000010. The results do not reflect FHWA's opinions.

\ifCLASSOPTIONcaptionsoff
  \newpage
\fi



%
\bibliographystyle{IEEEtran}
\bibliography{refs}

%

\begin{IEEEbiography}
[{\includegraphics[width=1in,height=1.25in,clip,keepaspectratio]{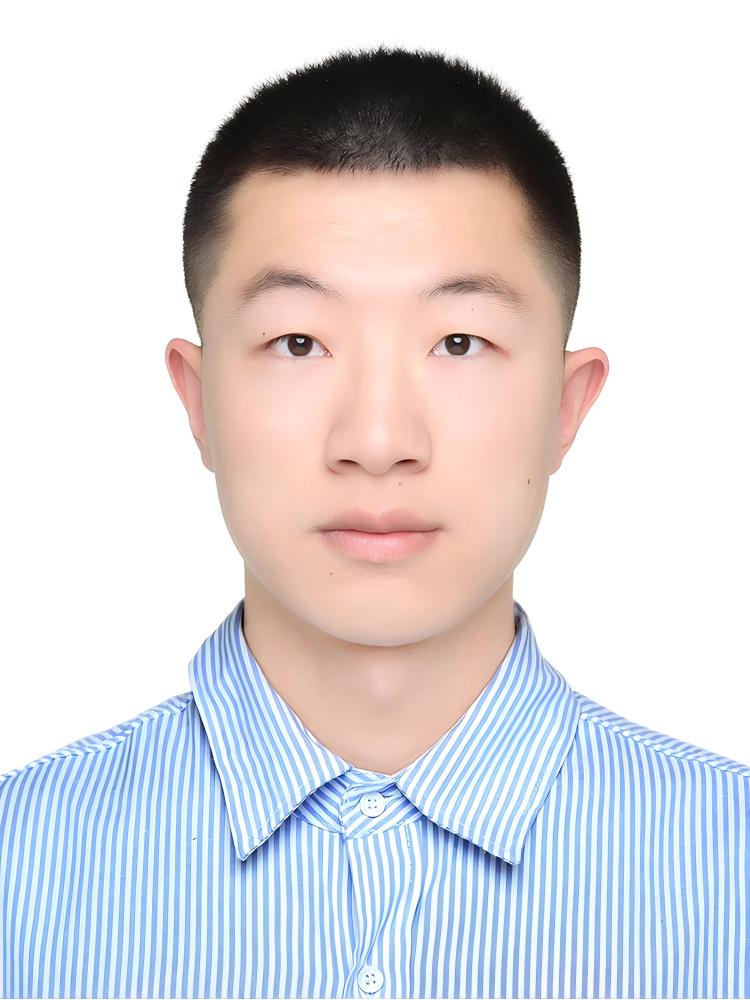}}]{Hao Zhang}
received a B.S. degree from Central South University, China, in 2020 and an M.S. degree in college of transportation engineering, Tongji University, China, in 2023. He is currently pursuing a Ph.D. degree in Zachry Department of Civil and Environmental Engineering, Texas A\&M University, College Station, TX, USA. 
His research interests include digital twin and traffic safety.
\end{IEEEbiography}
\begin{IEEEbiography}
[{\includegraphics[width=1in,height=1.25in,clip,keepaspectratio]{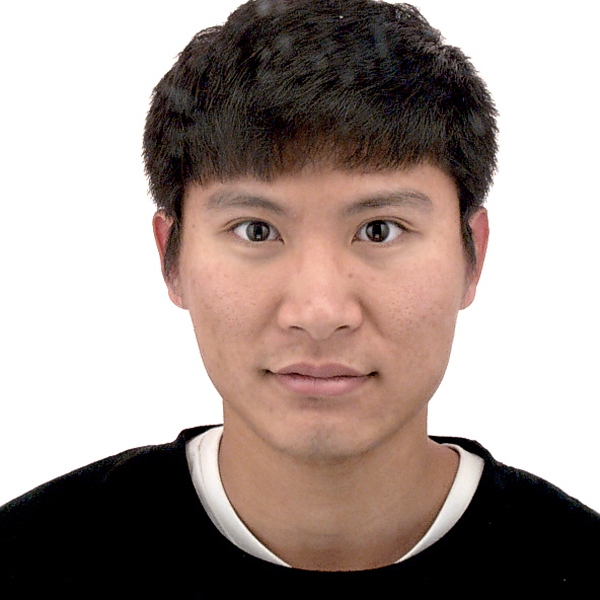}}]{Ximin Yue}
is a Ph.D. student at the Zachry Department of Civil and Environmental Engineering at Texas A\&M University. He holds a master of science degree in Industrial Engineering specializing in data science from Texas A\&M University in 2022. Additionally, he earned his bachelor’s degree in Industrial and System Engineering from Shanghai Maritime University and the University of New Haven, in 2020. Ximin’s research focuses on connected and autonomous vehicles, data science, and the application of deep reinforcement learning.
\end{IEEEbiography}

\begin{IEEEbiography}[{\includegraphics[width=1in,height=1.25in,clip,keepaspectratio]{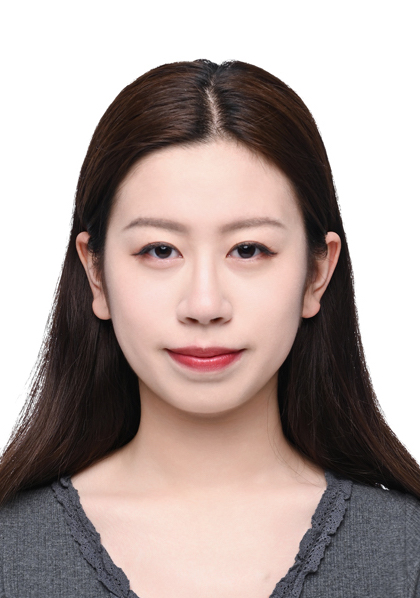}}]{Kexin Tian}
(Student Member, IEEE) serves as a research assistant at Texas A\&M University. She received her M.S. degree in Civil and Environmental Engineering from University of Wisconsin-Madison, WI, USA, in 2024, and the B.S. degrees in Computer Science and Mathematics from the University of Wisconsin-Madison, in 2022. She is currently pursuing her Ph.D. degree in Civil and Environmental Engineering with Texas A\&M University, College Station, TX, USA. Her main research directions are traffic flow prediction, intelligent transportation systems, and machine learning.
\end{IEEEbiography}

\begin{IEEEbiography}[{\includegraphics[width=1in,height=1.25in,clip,keepaspectratio]{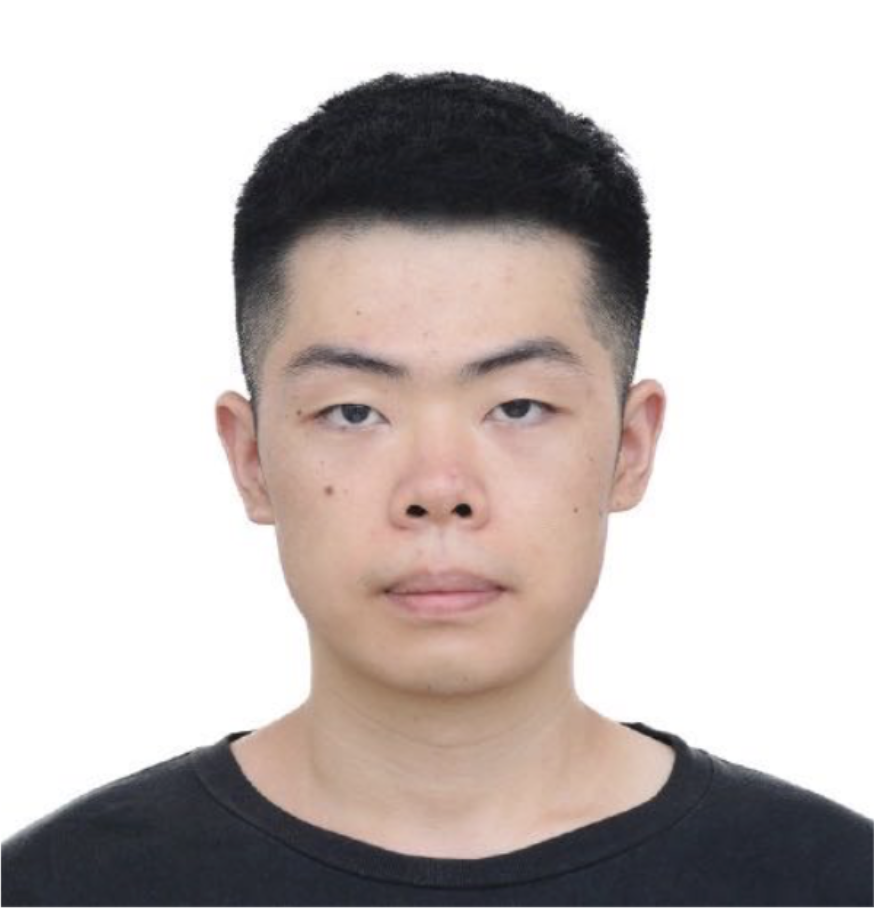}}]{Sixu Li}
(Student Member, IEEE) received a Bachelor's degree in Engineering Mechanics from Hunan University, Changsha, China, in 2020, and a Master's degree in Mechanical Engineering from UC Berkeley, Berkeley, CA, USA, in 2022. He is currently pursuing a Ph.D. degree in Civil and Environmental Engineering at Texas A\&M University, College Station, TX, USA.
His current research interests include dynamics and control, MPC, and optimization for autonomous driving, robotics, and  intelligent transportation systems.
\end{IEEEbiography}

\begin{IEEEbiography}[{\includegraphics[width=1in,height=1.25in,clip,keepaspectratio]{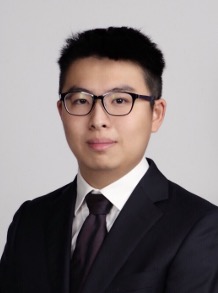}}]{Keshu Wu} is a postdoctoral research associate at Texas A\&M University. He receives his Ph.D. in Civil and Environmental Engineering from the University of Wisconsin-Madison in 2024. He also holds an M.S. degree in Civil and Environmental Engineering from Carnegie Mellon University in 2018 and an M.S. degree in Computer Sciences from the University of Wisconsin-Madison in 2022. He completed his B.S. in Civil Engineering at Southeast University in Nanjing, China in 2017. His research interests include the application and innovation of artificial intelligence and deep learning techniques in connected automated driving, intelligent transportation systems, and digital twin.
\end{IEEEbiography}
\begin{IEEEbiography}
[{\includegraphics[width=1in,height=1.25in,clip,keepaspectratio]{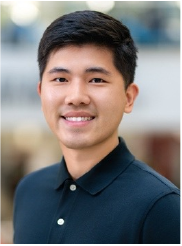}}]{Zihao Li} received a B.S. degree in Civil Engineering from Northeastern University, Shenyang, China, in 2017, an M.S. degree in Engineering Physics from Tsinghua University, Beijing, China, in 2020, and a Ph.D. in Civil Engineering from Texas A\&M University, College Station, TX, USA in 2024. His research interests center on the integration of data-driven and physical modeling in transportation systems, including connected automated vehicle systems, freight systems, transportation resilience, and traffic safety.
\end{IEEEbiography}
\begin{IEEEbiography}
[{\includegraphics[width=1in,height=1.25in,clip,keepaspectratio]{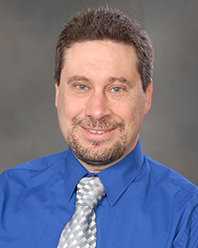}}]{Dominique Lord} received the B.Eng. degree in civil engineering from McGill University and the M.A.Sc. and Ph.D. degrees in civil engineering from the University of Toronto. He is currently a Professor and an A.P. and Florence Wiley Faculty Fellow with the Zachry Department of Civil Engineering, Texas A\&M University. His work focuses on conducting fundamental research in highway safety and crash data analyses.
\end{IEEEbiography}

\begin{IEEEbiography}
[{\includegraphics[width=1in,height=1.25in,clip,keepaspectratio]{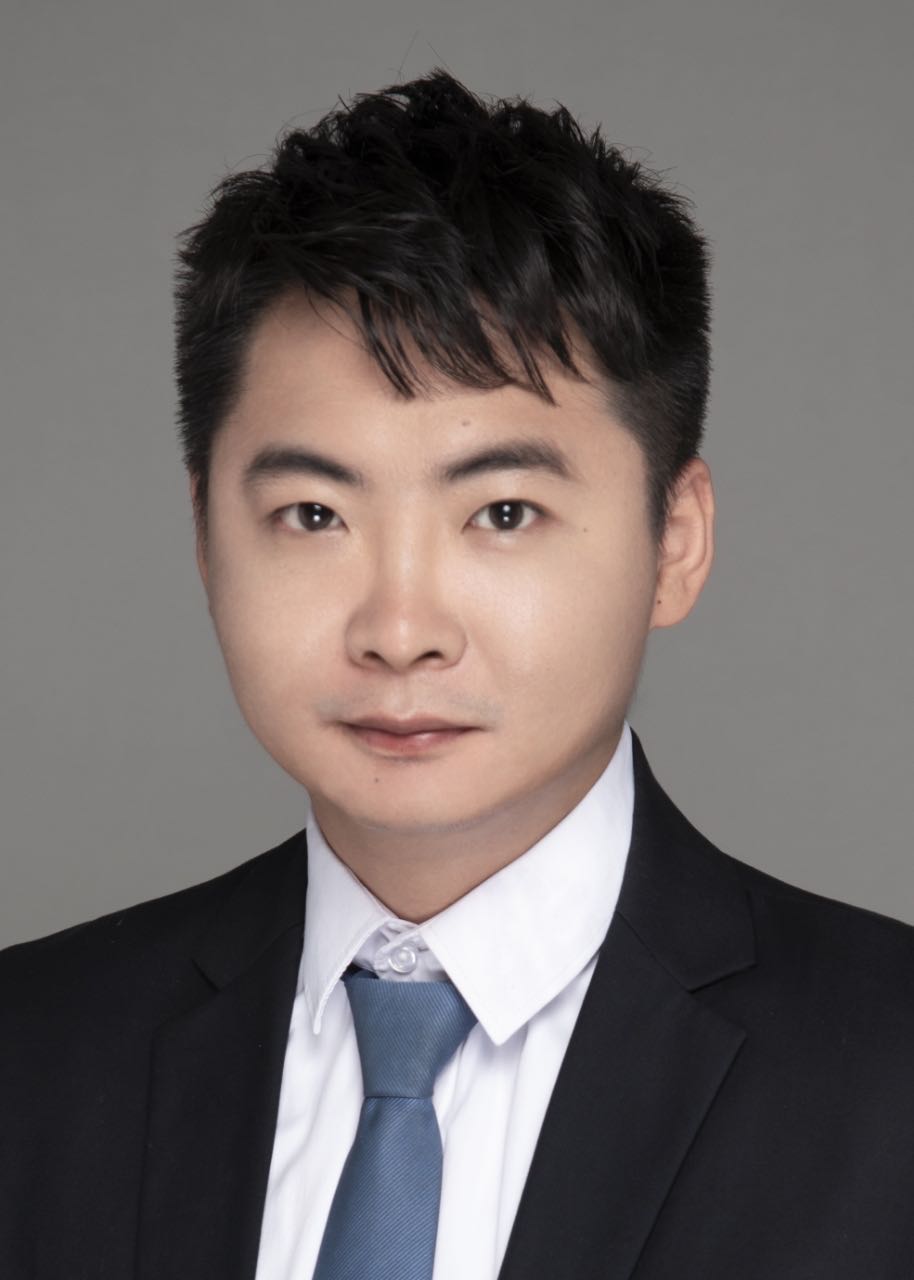}}]{Yang Zhou}
(Member, IEEE) received the Ph.D. degree in Civil and Environmental Engineering from University of Wisconsin Madison, WI, USA, in 2019, and the M.S. degree in Civil and Environmental Engineering from  University of Illinois at Urbana-Champaign, Champaign, IL, USA, in 2015. He is currently an assistant professor in the Zachry Department of Civil and Environmental Engineering, and Career Initiation Fellow in the Institute of Data Science, Texas A\&M University. Before joining Texas A\&M, he was a postdoctoral researcher in civil engineering at the University of Wisconsin Madison, WI, USA. He is currently a member of the TRB traffic flow theory CAV subcommittee, network modeling CAV subcommittee, and American Society of Civil Engineering TDI-AI committee. His main research directions are connected automated vehicles robust control, interconnected system stability analysis, traffic big data analysis, and microscopic traffic flow modeling.
\end{IEEEbiography}




\end{document}